\documentclass{article}
\pdfoutput=1

     \PassOptionsToPackage{numbers, compress}{natbib}

     \usepackage[preprint]{neurips_2019}

\usepackage[utf8]{inputenc} 
\usepackage[T1]{fontenc}    
\usepackage{hyperref}       
\usepackage{url}            
\usepackage{booktabs}       
\usepackage{amsfonts}       
\usepackage{nicefrac}       
\usepackage{microtype}      

\usepackage{ogonek}

\usepackage{graphicx}
\usepackage{subfigure}
\usepackage{algorithm}
\usepackage{algorithmic}
\usepackage{authblk}
\usepackage{comment}
\usepackage{amsmath}

\newcommand{\Eqref}[1]{Eq. \eqref{#1}}

\renewcommand{\algorithmiccomment}[1]{\bgroup\hfill $\triangleright$~#1\egroup}

\title{Gradient Noise Convolution (GNC): Smoothing Loss Function for Distributed Large-Batch SGD}

\renewcommand\Authands{, }
\author[1]{\textbf{Kosuke Haruki}}
\author[2,3]{\textbf{Taiji Suzuki}}
\author[1]{\textbf{Yohei Hamakawa}}
\author[1]{\textbf{Takeshi Toda}}
\author[1]{\authorcr \textbf{Ryuji Sakai}}
\author[1]{\textbf{Masahiro Ozawa}}
\author[1]{\textbf{Mitsuhiro Kimura}}
\affil[1]{Corporate Research and Development Center, Toshiba Corporation, Kawasaki, Japan}
\affil[2]{Graduate School of Information Science and Technology, The University of Tokyo, Japan}
\affil[3]{Center for Advanced Integrated Intelligence Research, RIKEN, Tokyo, Japan}
\affil[1]{\texttt{\{kosuke.haruki, yohei.hamakawa, takeshi1.toda, ryuji.sakai, masahiro3.ozawa, mitsuhiro4.kimura\}@toshiba.co.jp}}
\affil[2]{\texttt{taiji@mist.i.u-tokyo.ac.jp}}

\begin{document}

\maketitle

\begin{abstract}
Large-batch stochastic gradient descent (SGD) is widely used for training in distributed deep learning because of its training-time efficiency,
however, extremely large-batch SGD leads to poor generalization and easily converges to sharp minima,
which prevents na\"{\i}ve large-scale data-parallel SGD (DP-SGD) from converging to good minima.
To overcome this difficulty, we propose \emph{gradient noise convolution} (GNC),
which effectively smooths sharper minima of the loss function.
For DP-SGD, GNC utilizes so-called \emph{gradient noise},
which is induced by stochastic gradient variation and convolved to the loss function as a smoothing effect.
GNC computation can be performed by simply computing the stochastic gradient on each parallel worker and merging them,
and is therefore extremely easy to implement.
Due to convolving with the gradient noise, which tends to spread along a sharper direction of the loss function,
GNC can effectively smooth sharp minima and achieve better generalization, whereas isotropic random noise cannot.
We empirically show this effect by comparing GNC with isotropic random noise,
and show that it achieves state-of-the-art generalization performance for large-scale deep neural network optimization.
\end{abstract}

\section{Introduction}
\label{introduction}

Needs for faster training methods for deep learning are increasing with the availability of increasingly large datasets.
Distributed learning methods such as data-parallel stochastic gradient descent (DP-SGD),
which divides a large-batch into small-batches among parallel workers
and computes gradients on each worker, are widely used to accelerate the speed of training.
However, an open problem is that increasing the number of parallel workers increases DP-SGD large-batch sizes,
causing poor generalization (\citet{Goyal17}).
Various arguments have been made to explain this phenomenon.

One of the most convincing explanations, known as the sharp/flat hypothesis of the loss function,
is that large-batch SGD tends to converge to sharp minima.
While small-batch SGD introduces noisy behavior that allows escaping from sharp minima,
large-batch SGD has less randomness, resulting in the loss of this benefit.
From this observation, adding noise to gradients (or intermediate updates) may promote escaping from sharp minima,
while increasing batch sizes weaken such benefits (\citet{KeskarMNST16}).

Another convincing explanation along the lines of this hypothesis is that
the SGD training process has a smoothing effect on the loss function.
More specifically, SGD can be viewed as convolving stochastic noise
on the full-batch gradient in every training iteration,
eventually smoothing the loss function as the expectation (\citet{Kleinberg18}).
Larger batch sizes decrease stochastic noise,
resulting in the loss of this smoothing effect, and updates are therefore likely to be captured in bad local minima.

\begin{figure}[h]
  \begin{center}
  \centerline{\includegraphics[width=5in]{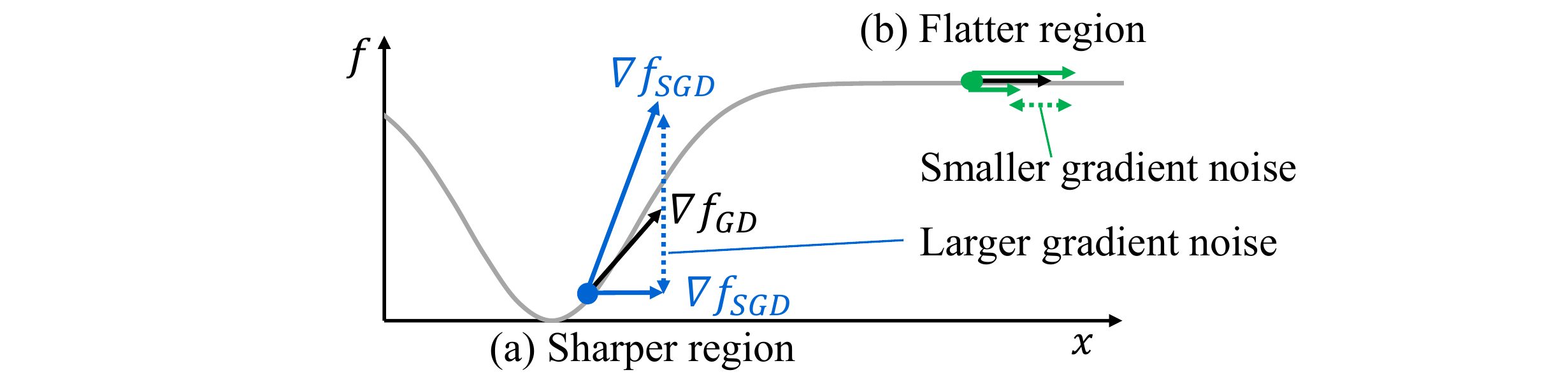}}
  \caption{
    Relation between curvature and gradient noise.
    (a) In sharper regions, stochastic gradients tend to vary
    because of unstableness of the loss landscape,
    and (b) vice versa in flatter regions.
  }
  \label{fig:noise_in_sharper_and_flatter_r2}
  \end{center}
  \vskip -0.3in
\end{figure}

Adding noise to the gradient in each update is a promising approach.
However, what kind of noise to add is not a trivial question.
One na\"{\i}ve method is to add isotropic noise, such as Gaussian noise with isotropic covariance (\citet{Wen18}), 
which would lead the solution toward an appropriate direction to some extent.
Indeed, \citet{Chaudhari17,Sagun17} experimentally showed
that the landscape of the loss function is nearly flat, with only a few sharp (steep) directions at each point.
Furthermore, those sharp directions are strongly aligned with the gradient variance.
In other words,
curvature of the loss function strongly correlates with covariance of the stochastic gradient.
Hence, variance of the stochastic gradient is strongly anisotropic (\citet{Daneshmand18,Sagun17}).
As a geometric intuition of this correlation,
Fig.~\ref{fig:noise_in_sharper_and_flatter_r2} shows the relation between curvature and gradient noise.
This observation partially explains the practical success of SGD,
whereas the artificial noise addition with isotropic variance
does not necessarily perform well in practice (\citet{Daneshmand18,Zhanxing18}).

We propose \emph{gradient noise convolution} (GNC) as a method for extremely large-scale DP-SGD
that utilizes {\it anisotropic} gradient noises.
Instead of adding artificially generated isotropic random noise,
we add noise naturally induced by the randomness of the stochastic gradient.
By doing so,
we can effectively avoid sharp minima and smooth sharp directions while reducing exploration in unproductive directions.
In summary, our proposed method has the following favorable properties:
\begin{enumerate}
\renewcommand{\labelenumi}{(\arabic{enumi})}
  \item high adaptivity for detecting sharp directions of the loss function
    by restoring gradient noise induced by the stochastic gradient but lost in extremely large-batch DP-SGD,
  \item easy implementation and computationally efficient smoothing of the loss function, and
  \item state-of-the-art generalization performance on an ImageNet-1K dataset
    with improved accuracy over existing methods for same numbers of training epochs.
\end{enumerate}

\paragraph{Relation to existing work}

\citet{Wen18} showed that convolution of isotropic random noise
can smooth the loss function to some extent.
However, this method does not selectively detect sharp directions, because the noise is isotropic.
In contrast,
we empirically show that our gradient noise method can more effectively detect sharp directions.

The most popular approach for detecting sharp directions is to utilize second-order information
such as the Hessian or Fisher information matrix.
While there are several theoretical and toy problem analyses of correlations
between covariance matrices of stochastic gradients and Hessian or Fisher matrices (\citet{Sagun17,jastrzebski18,Zhanxing18,YemingWen19}),
no computationally tractable proposals make use of curvature information and work in general settings.
Although \citet{Osawa18} demonstrated ImageNet training by the second-order method in practical times,
it is unclear whether the second-order method is truly computationally efficient
over standard SGD-based methods for datasets other than ImageNet.

Our approach for detecting sharper directions is simple and efficient:
our method utilizes only the first-order gradient
and obtains noise simply by computing differences between large-batch
and small-batch gradients computed on each parallel worker.
Generally, larger batch sizes ensure better estimations for the full gradient, but
decreasing gradient noise leads to poor generalization (\citet{Goyal17,Soatto17}).
In contrast, GNC can convolve multiple noises per iteration,
simply by adding gradient noises to calculations of the gradients on parallel workers in DP-SGD.
To summarize, GNC is the first tractable algorithm that is Hessian-free and
curvature adaptive for smoothing the loss function in large-scale DP-SGD training.
We also empirically show the robustness and effectiveness of GNC by applying it to
image classification benchmark datasets, namely CIFAR10, CIFAR100, and ImageNet-1K.

\section{Gradient Noise Convolution}
\label{sec:GNC}

We introduce a formalization of the GNC algorithm.
In the first section, we describe how GNC convolves multiple noises.
We also indicate similarity between the SGD update rule as noise convolution and GNC convolution.
Next, we show how GNC easily calculates gradient noises in extremely large-scale DP-SGD.
Finally, we define the GNC algorithm in detail.

\begin{figure}[h]
  \begin{center}
  \centerline{\includegraphics[width=6in]{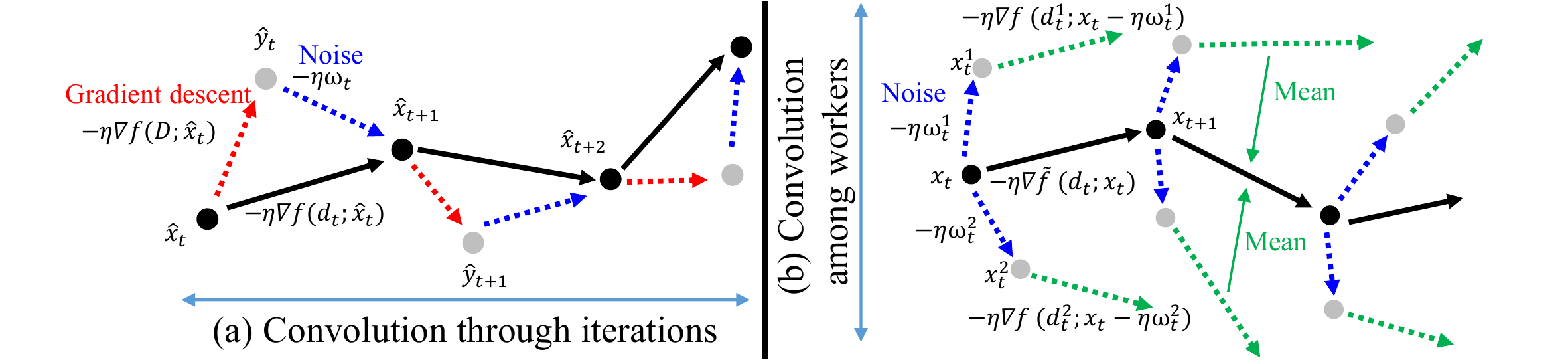}}
  \caption{
    (a){\bf Convolution through iterations}:
    The SGD update rule can be viewed as the GD update rule with noise convolution \emph{through} training iterations.
    (b){\bf Convolution among workers}:
    The GNC update rule can be viewed as the DP-SGD update rule with noise convolution \emph{among} parallel workers.
  }
  \label{fig:convolution_among_workers_and_iterations}
  \end{center}
  \vskip -0.3in
\end{figure}

\subsection{Convolution Through SGD Iterations}
\label{subsec:convolution_through_iterations}

Consider the SGD update rule for a loss function $f$ of a deep neural network.
Let $D = (z_i)_{i=1}^n$ denote the training dataset, $d_t \subset D$ be a mini-batch sampled from $D$,
$\hat{x}_t \in \mathbb{R}^{l}$ be a learnable parameter of $f$ at the iteration $t$,
and ${\eta}$ be the learning rate.
The SGD update rule is represented as
$\hat{x}_{t+1} = \hat{x}_{t} - {\eta} \nabla f(d_t;\hat{x}_{t})$ 
where $ \nabla f(d;x) := \frac{1}{|d|} \sum_{z \in d} \nabla f(z;x)$ for a mini-batch $d \subset D$
and $\nabla$ is the gradient with respect to the parameter $x$.
The stochastic gradient can be viewed as an unbiased estimator of full-gradient, so
the mini-batch gradient can be decomposed as
$\nabla f(d_t;\hat{x}_{t}) = \nabla f(D;\hat{x}_{t}) + {\hat{\omega}}_t$.
Here, we call the quantity ${\hat{\omega}}_t (:= \nabla f(d_t;\hat{x}_{t}) -\nabla f(D;\hat{x}_{t}))$ \emph{gradient noise}.
To describe the relation between the full gradient descent (GD) and SGD,
we introduce $\hat{y}_{t} := \hat{x}_{t} - {\eta} \nabla f(D;\hat{x}_{t})$,
where $\hat{y}_{t}$ is reached by the GD update rule.
From the equation $\hat{x}_{t+1} = \hat{y}_t - {\eta} {\hat{\omega}}_t$,
the SGD update rule can be rewritten as 
$\hat{y}_{t+1} = \hat{y}_{t} - {\eta} {\hat{\omega}}_t - {\eta} \nabla f(D;\hat{y}_{t} - {\eta} {\hat{\omega}}_t)$,
as Fig.~\ref{fig:convolution_among_workers_and_iterations}(a) shows.
Thus, the expectation through ${\hat{\omega}}_t$ leads to
    \begin{equation}
      {\mathbb{E}}_{{\hat{\omega}}_t} [\hat{y}_{t+1}]
      = \hat{y}_{t} - {\eta} \nabla {\mathbb{E}}_{{\hat{\omega}}_t} [f(D;\hat{y}_{t} - {\eta} {\hat{\omega}}_t)],
    \label{eq:alternative}
    \end{equation}
which shows SGD can be seen as GD with noise convolution (\citet{Kleinberg18}).
We call
$\nabla {\mathbb{E}}_{{\hat{\omega}}_t} [f(D;\hat{y}_{t} - {\eta} {\hat{\omega}}_t)]$
in \Eqref{eq:alternative} \emph{convolution through iteration}.

\subsection{Convolution Among Workers and Gradient Noise in GNC}
\label{subsec:convolution_among_workers}

Consider the DP-SGD update rule with $M$ parallel workers.
The large-batch gradient $\nabla f(d_{t};\hat{x}_{t})$ is 
the average of small-batch gradients $\nabla f(d_{t}^{i};\hat{x}_{t})$, computed as
    \begin{equation}
      \nabla f(d_{t};\hat{x}_{t}) = \frac{1}{M} \sum_{i=1}^{M} \nabla f(d_{t}^{i};\hat{x}_{t}),
    \label{eq:dp_sgd}
    \end{equation}
where $d_{t}^{i}$ is a small-batch obtained by splitting $d_t$ among the workers.
Inspired by the formulation of convolution through iteration in \Eqref{eq:alternative},
we modified the right side of \Eqref{eq:dp_sgd} to
define \emph{convolution among workers} as
    \begin{equation}
      \nabla \tilde{f}(d_t;x_{t}) =
        \frac{1}{M} \sum_{i=1}^{M} \nabla f(d_{t}^{i};x_{t} - {\eta} {\omega}_{t}^{i}),
    \label{eq:gradient_convolved_function_r2}
    \end{equation}
where ${\omega}_{t}^{i}$ denotes gradient noise defined below in \Eqref{eq:gradient_noise_definition_r2},
$x_{t} \in \mathbb{R}^{l}$ denotes a learnable parameter of $\tilde{f}$ at the iteration $t$,
and $\tilde{f}$ denotes the smoothed function
$\tilde{f}(d_t;x_{t}) := \frac{1}{M} \sum_{i=1}^{M} f(d_{t}^{i};x_{t} - {\eta} {\omega}_{t}^{i})$.
As Fig.~\ref{fig:convolution_among_workers_and_iterations}(b) shows,
the $i$th worker injects the noise ${\omega}_{t}^{i}$ to the weight $x_t$
and calculates the gradient $\nabla f(d_{t}^{i};x_{t} - {\eta} {\omega}_{t}^{i})$.
All gradients of the parallel workers are then averaged as the large-batch gradient $\nabla \tilde{f}(d_t;x_{t})$.
Generally, as the number $M$ of the workers are increased in DP-SGD,
the large-batch size is also increasing, leading to a better estimation of the full gradient.
Conversely, this decreases gradient noise, thus leading to poor generalization.
However, by adapting convolution among workers to DP-SGD,
the number of noise convolutions can be replenished $M$ times per iteration, thus retaining good generalization.
The noise term ${\omega}_{t}^{i}$ introduced in \Eqref{eq:gradient_convolved_function_r2} is calculated on each parallel worker and 
defined as
    \begin{equation}
      {\omega}_{t}^{i} :=
        \nabla f(d_{t-1}^{i};x_{t-1} - {\eta} {\omega}_{t-1}^{i}) - \nabla \tilde{f}(d_{t-1};x_{t-1}).
    \label{eq:gradient_noise_definition_r2}
    \end{equation}
When $M$ or the size of $d_{t-1}$ is sufficiently large,
$\nabla f(d_{t-1};x_{t-1})$ in \Eqref{eq:gradient_noise_definition_r2} can be approximately seen as the full-batch gradient.
Hence, ${\omega}_{t}^{i}$ can well estimate the gradient noise.
Note that, ${\omega}_{t}^{i}$ is the unbiased noise with respect to $i$
because of $\sum_{i=1}^{M} {\omega}_{t}^{i} = 0$,
which is derived from \Eqref{eq:gradient_noise_definition_r2}.

\subsection{Random Noise Convolution}
\label{subsec:RNC}

As a counter part of our GNC method, 
we introduce \emph{random noise convolution} (RNC), which unlike GNC utilizes isotropic random noise.
Instead of using \Eqref{eq:gradient_noise_definition_r2}, 
we define ${\omega}_{t}^{i}$ as a uniformly distributed random vector
on the unit cube along each dimension.
Note that ${\omega}_{t}^{i}$ in RNC is also the unbiased noise with respect to $i$.
Sec.~\ref{subsec:comparison_gn_rn} presents a detailed comparison of GNC with RNC.


\subsection{Algorithm for Gradient Noise Convolution}
\label{subsec:algo_gnc}

Algorithm~\ref{alg:GNC} shows pseudo-code for GNC.
Note that, in addition to \Eqref{eq:gradient_convolved_function_r2},
we introduce a hyperparameter $\alpha$ to control the window size in the convolution.
Our GNC method only adds the two subtractions to the standard DP-SGD update rule.
Specifically, ``compute \emph{gradient noise}'' and ``compute weights with noise'' in Algorithm~\ref{alg:GNC}
are the only differences from standard DP-SGD.
Hence, GNC can be easily implemented and efficiently computed.

\begin{algorithm}[ht]
  \caption{Gradient noise convolution}
  \label{alg:GNC}
\begin{algorithmic}
  \STATE {\bfseries Input:} loss function $f$, weights $x$, training dataset $D$,
    learning rate $\eta$, \\
    number of workers $M$, total iterations $T$, noise coefficient $\alpha$
  \STATE {\bfseries Init:} $x_0$ is randomly initialized, ${\omega}_{0}^{i}$ is $0$
  \FOR{$t=1$ {\bfseries to} $T$}
    \STATE Choose mini-batch $d_t$ with size ($b \times M$) from $D$
    \FOR{$i=1$ {\bfseries to} $M$}
      \STATE Choose local mini-batch $d_{t}^{i}$ with size $b$ from $d_t$
      \IF{$t > 1$}
        \STATE
          ${\omega}_{t}^{i} \leftarrow g_{t-1}^{i} - \tilde{g}_{t-1}$
          \COMMENT{compute \emph{gradient noise}}
      \ENDIF
      \STATE
        $x_{t}^{i} \leftarrow x_{t} - \alpha {\eta}_t \omega_{t}^{i}$
        \COMMENT{compute weights with noise}
      \STATE
        $g_{t}^{i} \leftarrow \frac{1}{b} \sum_{j=1}^{b}
         \nabla f (d_{t}^{i,j};x_{t}^{i})$ where $d_t^{i}=(d_t^{i,1},\dots,d_t^{i,b})$
        \COMMENT{compute gradient on each worker}
    \ENDFOR
    \STATE
      $\tilde{g}_{t} \leftarrow \frac{1}{M} \sum_{i=1}^{M} g_{t}^{i}$
      \COMMENT{compute gradient for the convolved function}
    \STATE
      $x_{t+1} \leftarrow x_t - {\eta}_t \tilde{g}_{t}$
      \COMMENT{update weights}
  \ENDFOR
\end{algorithmic}
\end{algorithm}

\section{Empirical Analysis for Effectiveness of GNC}
\label{empirical_analysis}

In this section, we describe how GNC effectively smooths the loss function.
We empirically show performance related to
full-gradient predictiveness by the large-batch gradient,
ability to detect sharper directions from gradient noise,
and effectiveness of smoothing the loss function by GNC.
We also demonstrate the robustness of GNC with and without modern optimization techniques,
such as momentum (MM), weight decay (WD) and data augmentation (DA).

Note that, in all experiments, except for ImageNet, we completely eliminated randomness in the training process.
Random initial weights or mini-batch random sampling or non deterministic floating-point calculations
can hide true differences among different training methods.
Therefore, when benchmarking against DP-SGD, GNC, RNC, and so on,
we used the same initial random weight and the same random mini-batch sampling and deterministic mode for GPUs.
In the ImageNet case, we could not eliminate floating-point calculation randomness due to performing AllReduce,
which is a collective communication operation among multi-nodes.
For example, ``five runs'' means 
using five sets of random values including random initial weights and random sequences of mini-batch sampling.
We chose one of these five random sets and applied the same set to DP-SGD, GNC, and RNC for fair comparison.
We iterated this five times.
To ensure a fair and precise benchmark process, we selected no special random value sets.

\subsection{Similarity Between Full and Large-batch Gradients}
\label{subsec:similarity_fg_lb}

We first investigated whether the large-batch gradient truly approximates the full gradient.
As a baseline for this experiment,
we considered training a standard ResNet-32 architecture on CIFAR-100 by DP-SGD for 160 epochs.
We used a large-batch size of 8,192 comprising 256 parallel workers with a small-batch size of 32.
We adopted some learning rate scheduling techniques.
For example, we applied step decay at the 80th and 120th epochs, gradual warmup,
and the linear scaling rule (\citet{Goyal17}).
To stabilize extremely large batch training, we adopted batch normalization (BN), WD, MM, and DA.
See Appendix~\ref{detail_experimental_setup} for complete training settings.

\begin{figure}[ht]
  \vskip -0.2in
  \begin{minipage}{0.33\hsize}
    \begin{center}
      \def\subfigcapskip{0pt}
      \subfigure[Validation accuracy]{
        \includegraphics[width=1.8in]{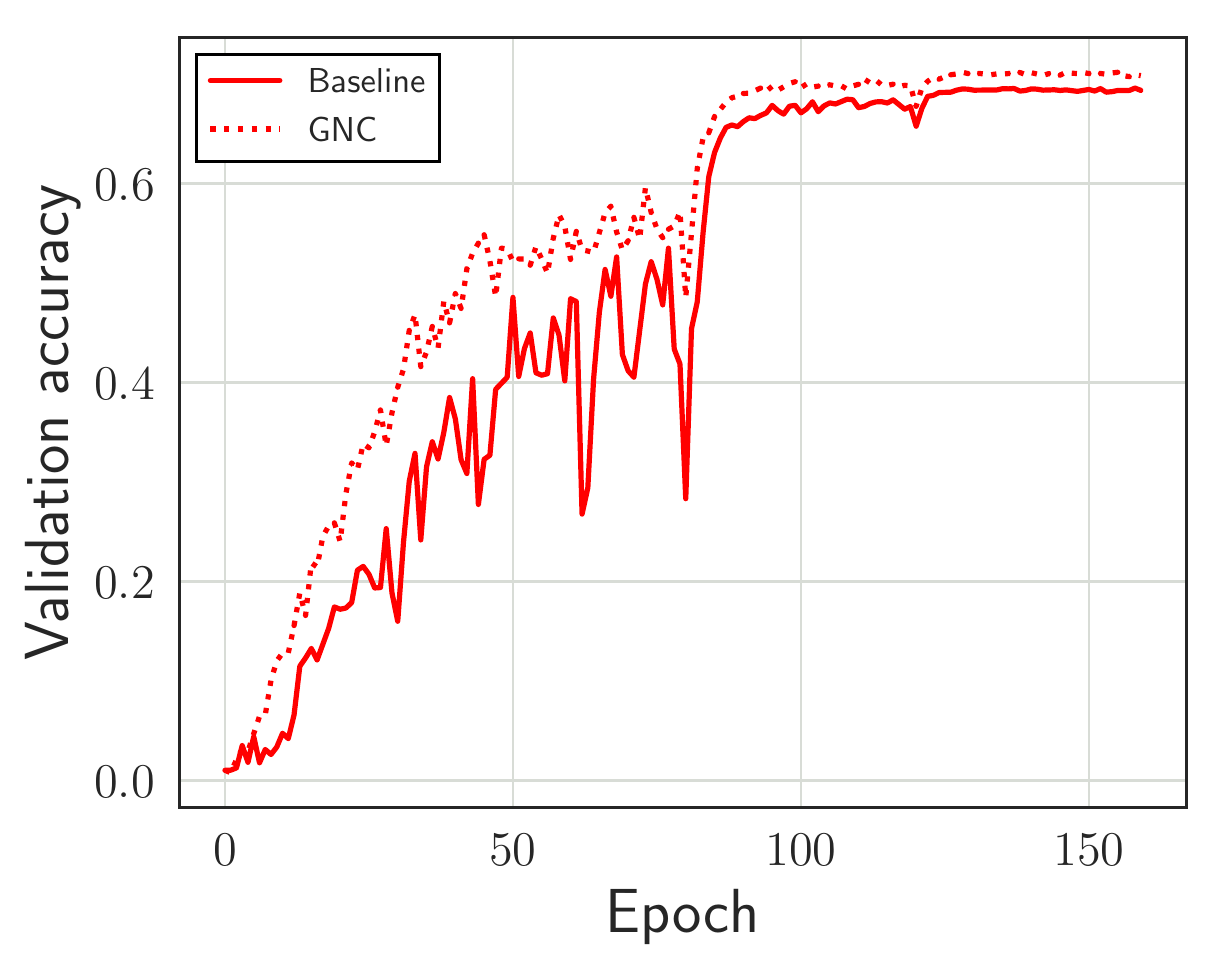}}
    \end{center}
  \end{minipage}
  \begin{minipage}{0.33\hsize}
    \begin{center}
      \def\subfigcapskip{0pt}
      \subfigure[FG similarity]{
        \includegraphics[width=1.8in]{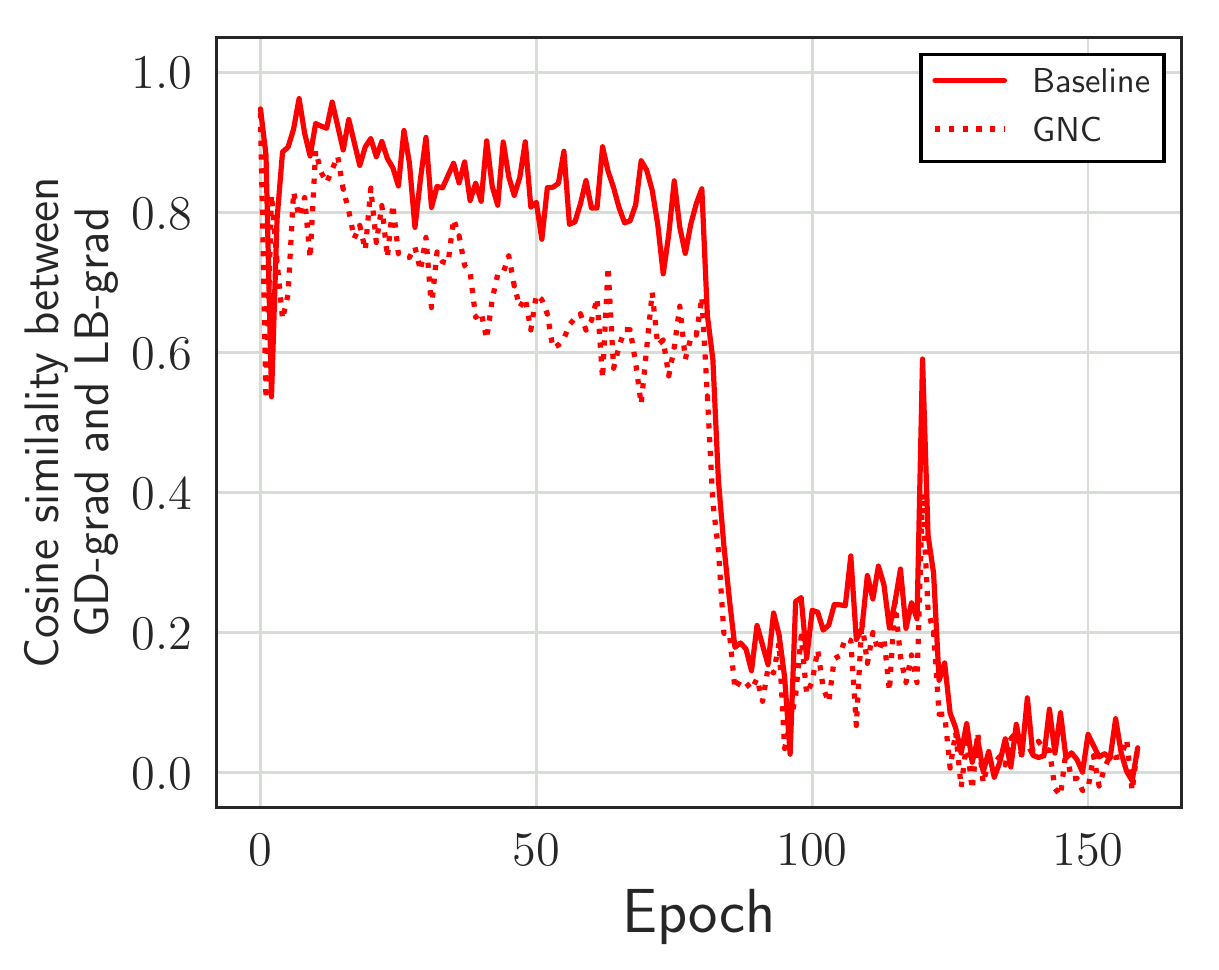}}
    \end{center}
  \end{minipage}
  \begin{minipage}{0.33\hsize}
    \begin{center}
      \def\subfigcapskip{0pt}
      \subfigure[FG similarity of SGD]{
        \includegraphics[width=1.8in]{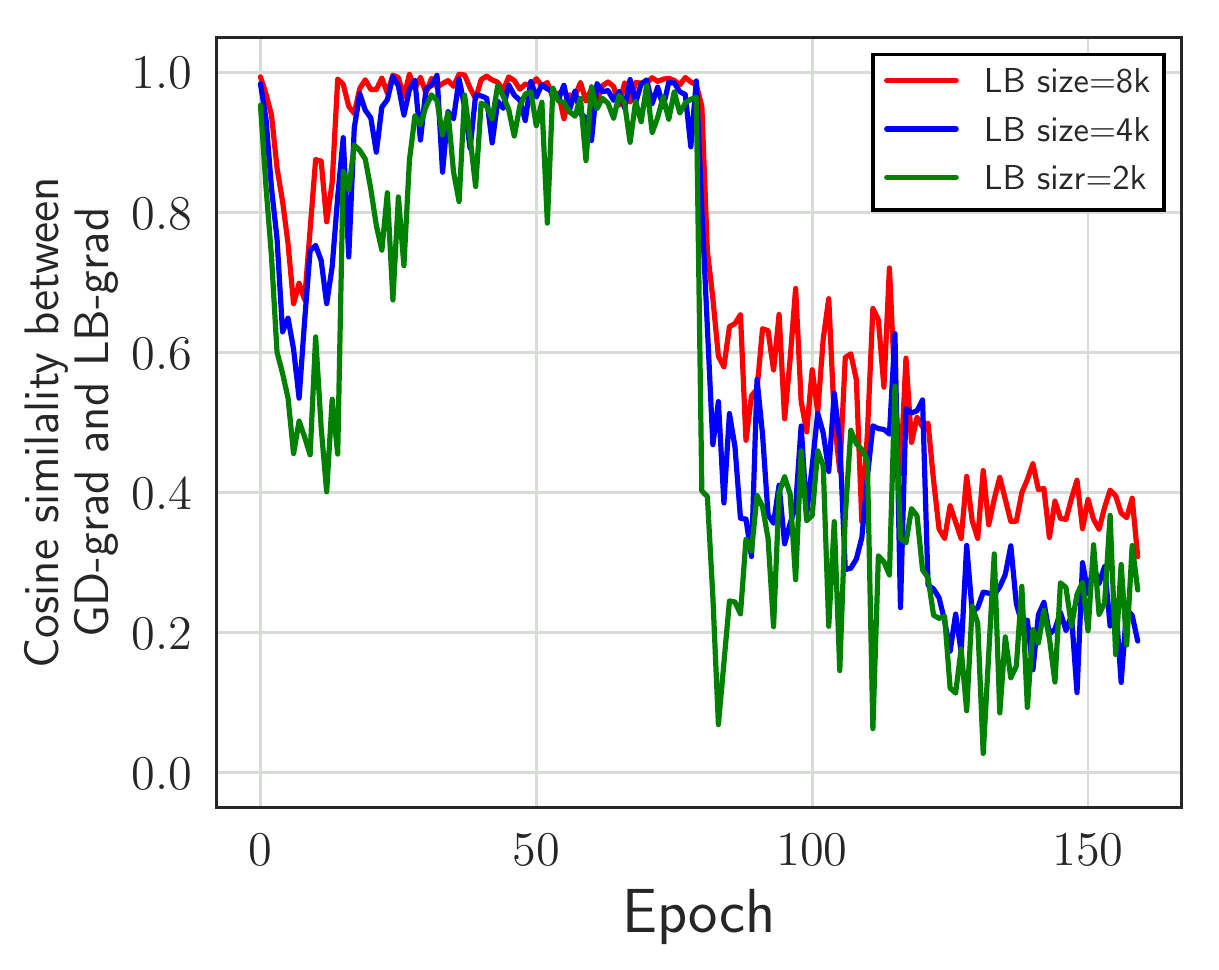}}
    \end{center}
  \end{minipage}
  \vskip -0.15in
  \caption{
    {\bf (a)}~Validation accuracy of training with and without GNC.
    Applying GNC improved validation accuracy.
    {\bf (b)}~Cosine similarity between the full gradient (FG) and large-batch gradient.
    The learning rate was step-decayed at the 80th and 120th epochs.
    Concurrently, cosine similarity significantly dropped.
    Note that to eliminate side-effects of noise convolution, the full gradient was calculated without GNC.
    In contrast, large-batch gradient of GNC indicates smoothed large-batch gradient defined in \Eqref{eq:gradient_convolved_function_r2}.
    {\bf (c)}~Cosine similarity between FG and large-batch gradient
    as calculated by 2K, 4K, and 8K batch-size training models.
    We trained these models by simple SGD (i.e., without BN, WD, DA, and MM from SGD optimization).
    Similarities of all models dropped as in (b),
    so the predominant factor behind the similarity drop was step decay.
    It is also interesting that a larger batch size mitigates similarity drops.
    In future work we will investigate in detail the mechanism behind this similarity drop,
    but an intuitive understanding is that (1)~a smaller step size can fall into a narrower region (\citet{jastrzebski18}),
    and (2)~this narrower region might be a complicated high-dimensional region.
    It also seems that the learning process of deep nets initially acquires simple features
    and gradually acquires complex features.
  }
  \label{fig:cifar100_full_gradient_similarity}
  \vskip -0.05in
\end{figure}

We first showed that GNC improves the validation accuracy of CIFAR-100 training (Fig.~\ref{fig:cifar100_full_gradient_similarity}(a)).
We also measured predictiveness of the full gradient by large-batch gradient (Fig.~\ref{fig:cifar100_full_gradient_similarity}(b)).
At early training stages, both the baseline and GNC models indicate good predictiveness.
Close observation shows that this is due to the faster GNC training process, as seen in Fig.~\ref{fig:cifar100_full_gradient_similarity}(a),
so GNC predictiveness was slightly worse than the baseline.
Predictiveness significantly dropped in later training stages, coinciding with the step decay of the learning rate.
We performed an ablation study to investigate this further, training only by learning rate scheduling without BN, WD, DA, and MM.
Fig.~\ref{fig:cifar100_full_gradient_similarity}(c) shows the results of this ablation study,
which indicate that step decay is a key factor behind the drop.
Further, larger batch sizes mitigate the drop (Fig.~\ref{fig:cifar100_full_gradient_similarity}(c)).

\subsection{Comparison of Gradient Versus Random Noise}
\label{subsec:comparison_gn_rn}

We investigated the effectiveness of GNC for detecting sharper regions
in comparison with RNC.
We first demonstrated anisotropy of GNC and RNC.
We define the condition number ${\kappa}_{t}$ for the covariance matrix of noise as,
\def\vector#1{\mbox{\boldmath $#1$}}
  ${\Sigma}_{t} := \frac{1}{M} \vector{{\Omega}}_{t} {{\vector{\Omega}}_{t}}^{\top} \in \mathbb{R}^{l \times l}$,
  ${\kappa}_{t} := \frac{{\lambda}_{\rm max}({\Sigma}_{t})}{{\lambda}_{\rm min^{*}}({\Sigma}_{t})}$,
where $M$ is the number of workers,
$\omega_{t}^{i}$ is the gradient noise computed on worker $i$ as in Sec.~\ref{sec:GNC},
$\overline{\omega}_{t} := \frac{1}{M} \sum_{i=1}^{M} \omega_{t}^{i} \in \mathbb{R}^{l}$,
and $\vector{\Omega}_{t} := [\omega_{t}^{1} - \overline{\omega}_{t}, \omega_{t}^{2} - \overline{\omega}_{t},
\cdots, \omega_{t}^{M} - \overline{\omega}_{t}] \in \mathbb{R}^{l \times M}$.
To derive ${\kappa}_{t}$, we computed the maximum eigenvalue of ${\Sigma}_{t}$ as
${\lambda}_{\rm max}({\Sigma}_{t})$, and its second-smallest eigenvalue as 
${\lambda}_{\rm min^{*}}({\Sigma}_{t})$.
Note that, due to $M \ll l$, we calculated eigenvalues of
$\frac{1}{M} {{\vector{\Omega}}_{t}}^{\top} \vector{{\Omega}}_{t} \in \mathbb{R}^{M \times M}$
instead of ${\Sigma}_{t}$ for computational efficiency.
From the nature of the condition number, a larger condition number indicates noise with higher anisotropy.

\begin{figure}[ht]
  \vskip -0.2in
  \begin{minipage}{0.33\hsize}
    \begin{center}
      \def\subfigcapskip{0pt}
      \subfigure[Noise covariance]{
        \includegraphics[width=1.8in]{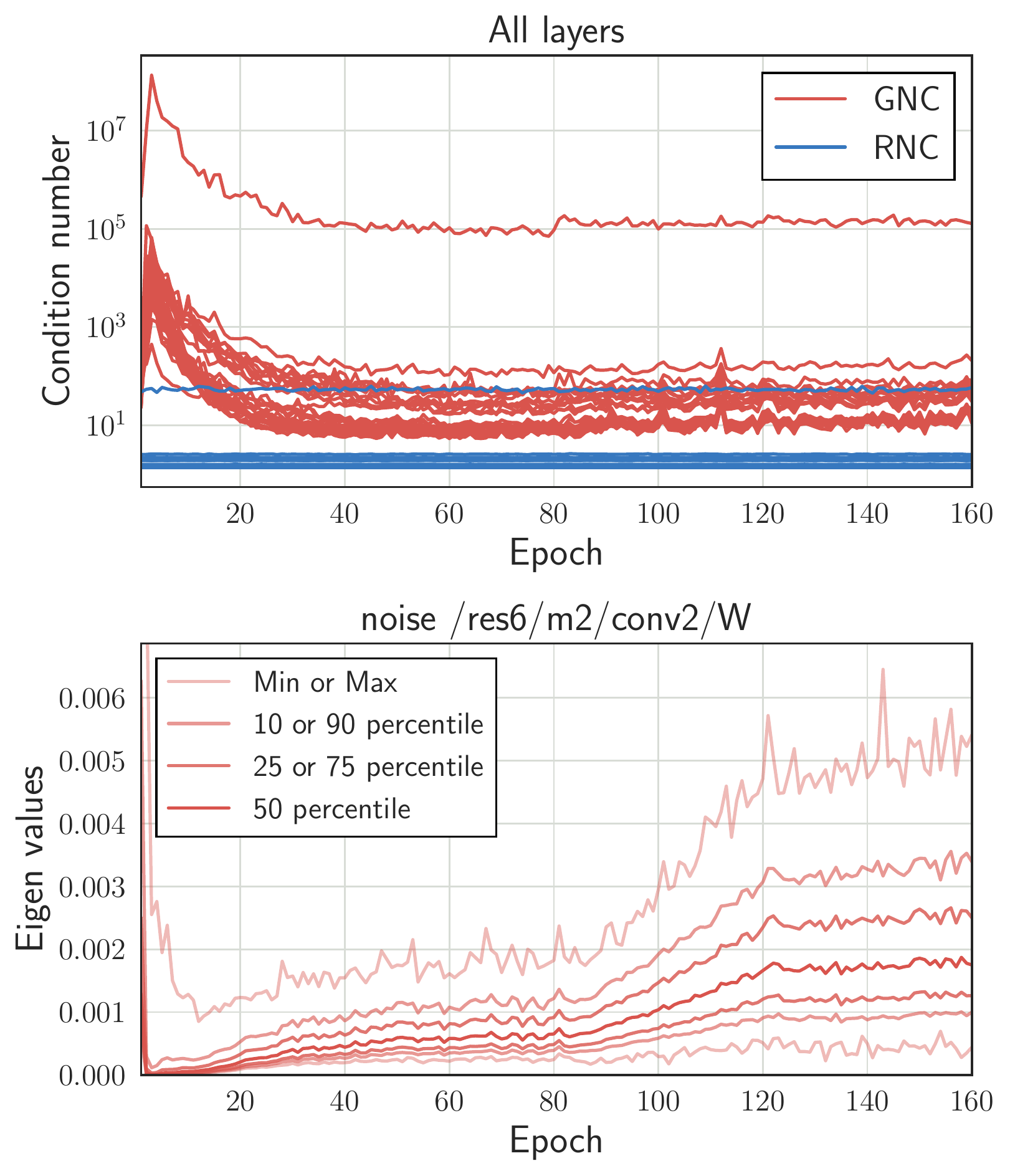}}
    \end{center}
  \end{minipage}
  \begin{minipage}{0.33\hsize}
    \begin{center}
      \def\subfigcapskip{0pt}
      \subfigure[Losses in 10 epochs]{
        \includegraphics[width=1.8in]{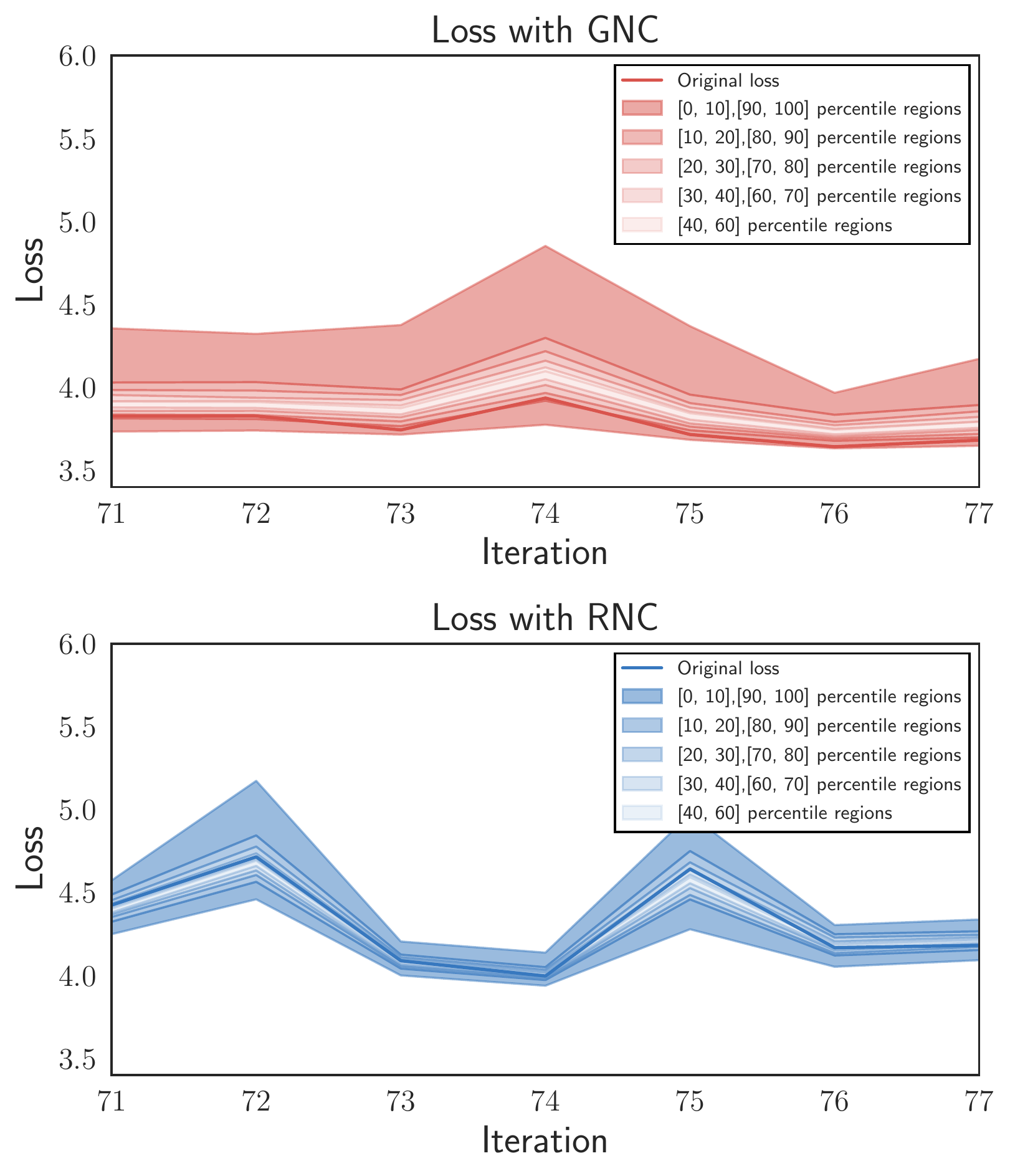}}
    \end{center}
  \end{minipage}
  \begin{minipage}{0.33\hsize}
    \begin{center}
      \def\subfigcapskip{0pt}
      \subfigure[Losses in 140 epochs]{
        \includegraphics[width=1.8in]{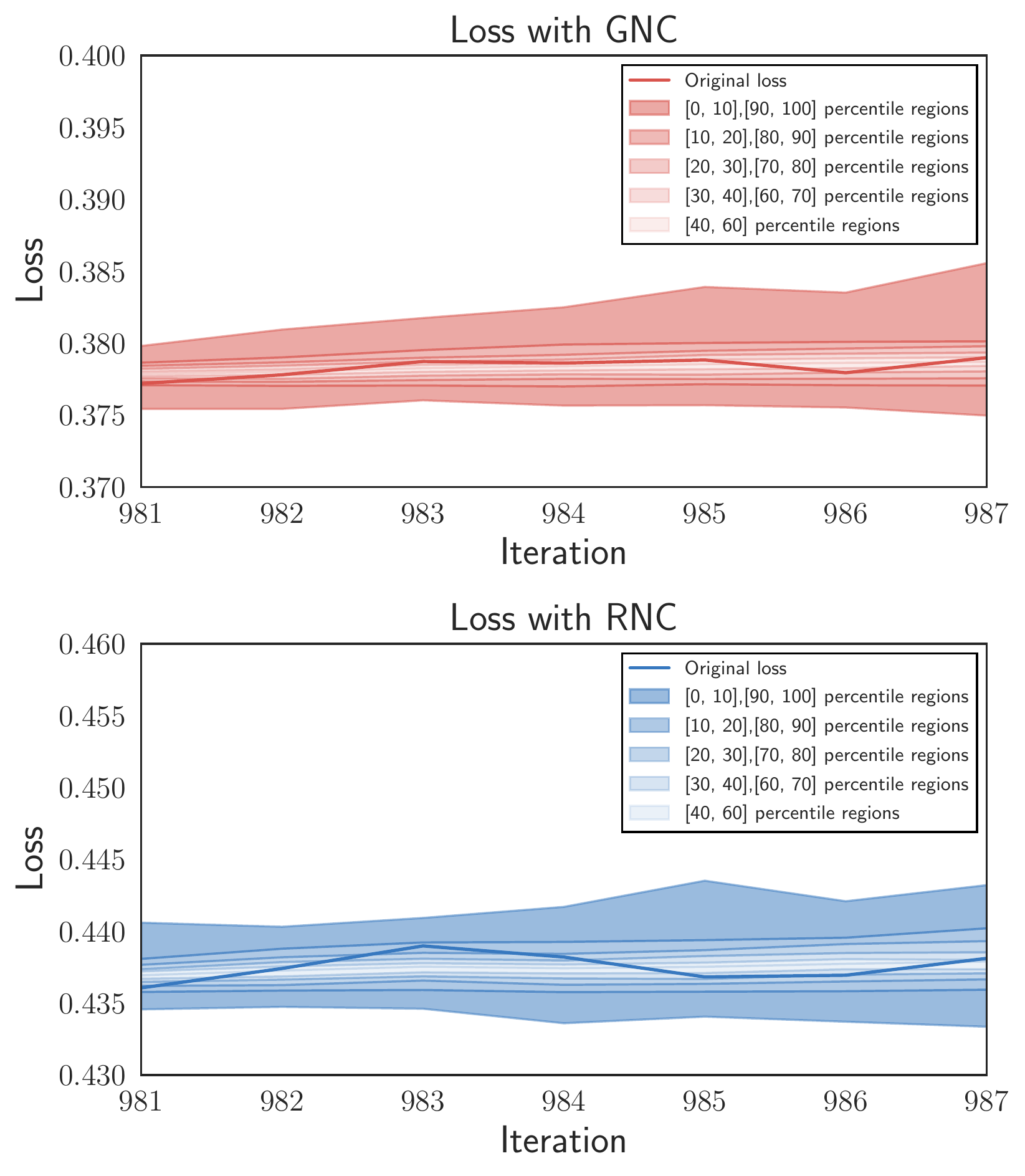}}
    \end{center}
  \end{minipage}
  \vskip -0.15in
  \caption{
    {\bf (a)~Upper:} Condition numbers for GNC and RNC noise covariance through the training process.
    We calculated condition numbers for each ResNet32 layer and plot lines layer-by-layer,
    so different lines indicate different layers.
    Condition numbers for GNC are larger than those for RNC, 
    so it is obvious that noise in GNC is highly anisotropic and noise in RNC is isotropic.
    {\bf (a)~Lower:} Eigenvalues of GNC noise covariance.
    We calculated eigenvalues of a selected ResNet32 layer, namely /res6/m2/conv2/W.
    Different lines are corresponding to the percentile magnitude of eigenvalues.
    Eigenvalues increase with training process,
    possibly indicating that the model is heading toward a complex region.
    This situation possibly correlates with a drop in predictiveness of the full gradient,
    as shown in Fig.~\ref{fig:cifar100_full_gradient_similarity}(c).
    {\bf (b), (c):}
    We measured losses of the models $x_{t}^{i}$ $(i=1,...,M)$ in 
    Fig.~\ref{fig:convolution_among_workers_and_iterations}(b).
    The bold line shows loss of $x_{t}$ through iteration $t$.
    The shaded region indicates losses of $x_{t}^{i}$ among workers $i$.
    If noise ${\omega}_{t}^{i}$ correctly heads towards a sharper direction,
    losses of $x_{t}^{i}$ will vary from the loss of $x_{t}$.
    {\bf (b):} Losses around the 10th epochs of GNC (upper) and RNC (lower). As seen in Sec.~\ref{subsec:similarity_fg_lb},
    due to the good predictiveness of the full gradient at the 10th epochs,
    GNC noise can correctly detect sharper directions, so losses widely vary.
    In contrast, because RNC noise is random and isotropic, losses vary less than in GNC.
    {\bf (c):} Losses around the 140th epochs.
    While GNC is still anisotropic, predictiveness of the full gradient decreases around the 140th epochs.
    Hence, GNC cannot detect the sharper directions,
    and losses do not vary much in both GNC and RNC.
  }
  \label{fig:cifar100_gradient_noise_vs_random_noise}
  \vskip -0.05in
\end{figure}

The upper panel in Fig.~\ref{fig:cifar100_gradient_noise_vs_random_noise}(a) shows that noise with GNC has larger condition numbers than with RNC,
meaning GNC noise is highly anisotropic.
We have thus established that GNC noise is highly anisotropic, but this does not necessarily mean this noise actually detects sharper directions.
To investigate their ability to detect sharper directions,
we measured the losses of models $x_{t}^{i}$ (Fig.~\ref{fig:convolution_among_workers_and_iterations}(b)),
which are perturbed models of $x_{t}$ toward noise directions.
Fig.~\ref{fig:cifar100_gradient_noise_vs_random_noise}(b) and (c) show the losses of these models.
Although GNC losses vary more widely than those with RNC in (b),
there are no significant differences in (c),
where full gradient predictiveness drops as described in Sec.~\ref{subsec:similarity_fg_lb}.
GNC can thus detect sharper directions, as long as the large-batch gradient correctly estimates the full gradient.

\subsection{GNC Effectiveness in Smoothing the Loss Function}
\label{subsec:gnc_smoothing_evidence}

This section directly shows the GCN smoothing effect.
As Sec.~\ref{subsec:comparison_gn_rn} showed, although GNC can detect sharper directions and average widely varying losses,
it remains unclear whether doing so effectively smooths the loss function.
To investigate this question, 
we introduce three indicators shown to be effective by \citet{Shibani18}:
loss stability (i.e., Lipschitzness), gradient predictiveness, and Lipschitzness of the gradient (i.e. $\beta$-smoothness).

\begin{figure}[ht]
  \vskip -0.18in
  \begin{minipage}{0.33\hsize}
    \begin{center}
      \def\subfigcapskip{0pt}
      \subfigure[Loss landscape]{
        \includegraphics[width=1.8in]{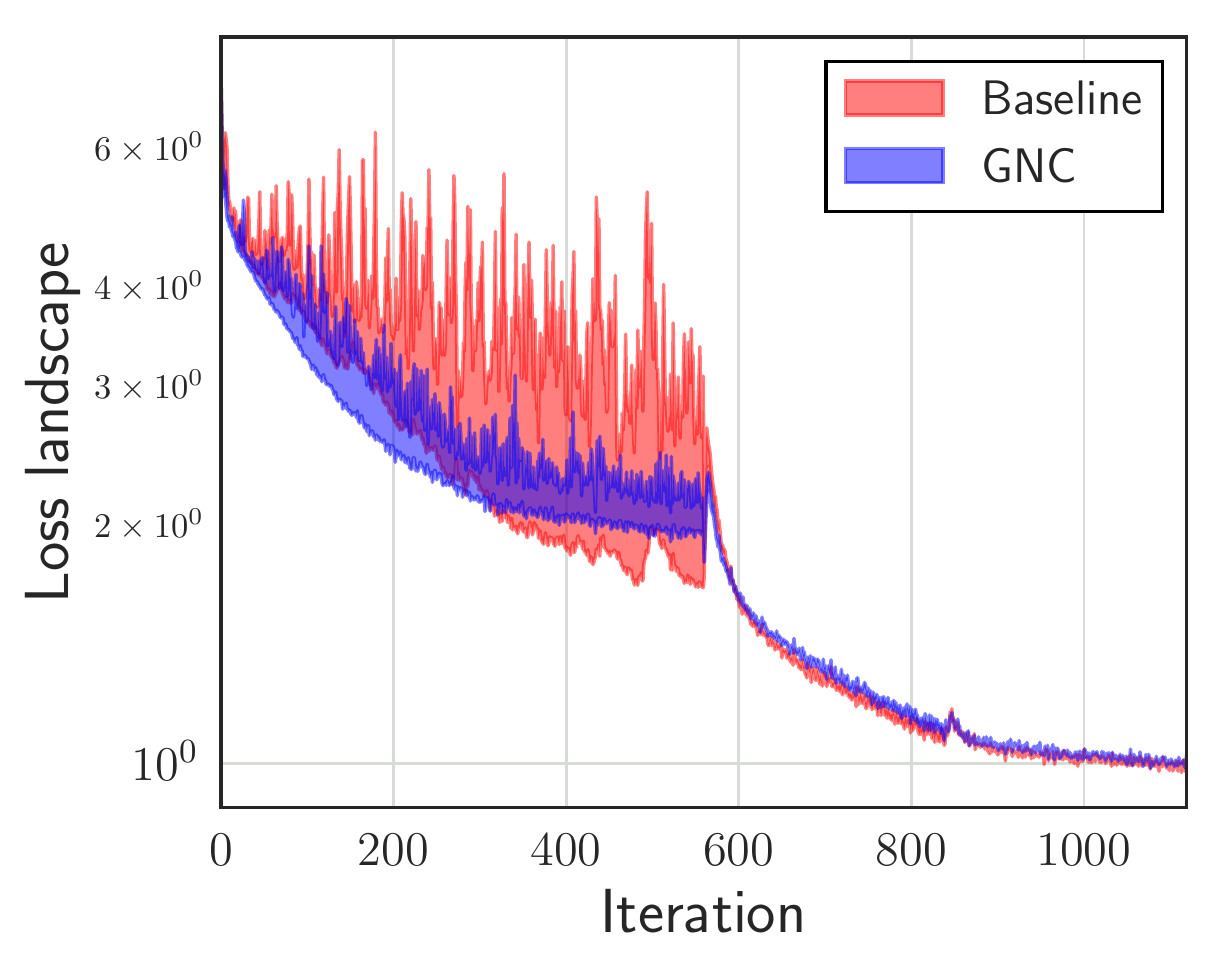}}
    \end{center}
  \end{minipage}
  \begin{minipage}{0.33\hsize}
    \begin{center}
      \def\subfigcapskip{0pt}
      \subfigure[Gradient predictiveness]{
        \includegraphics[width=1.8in]{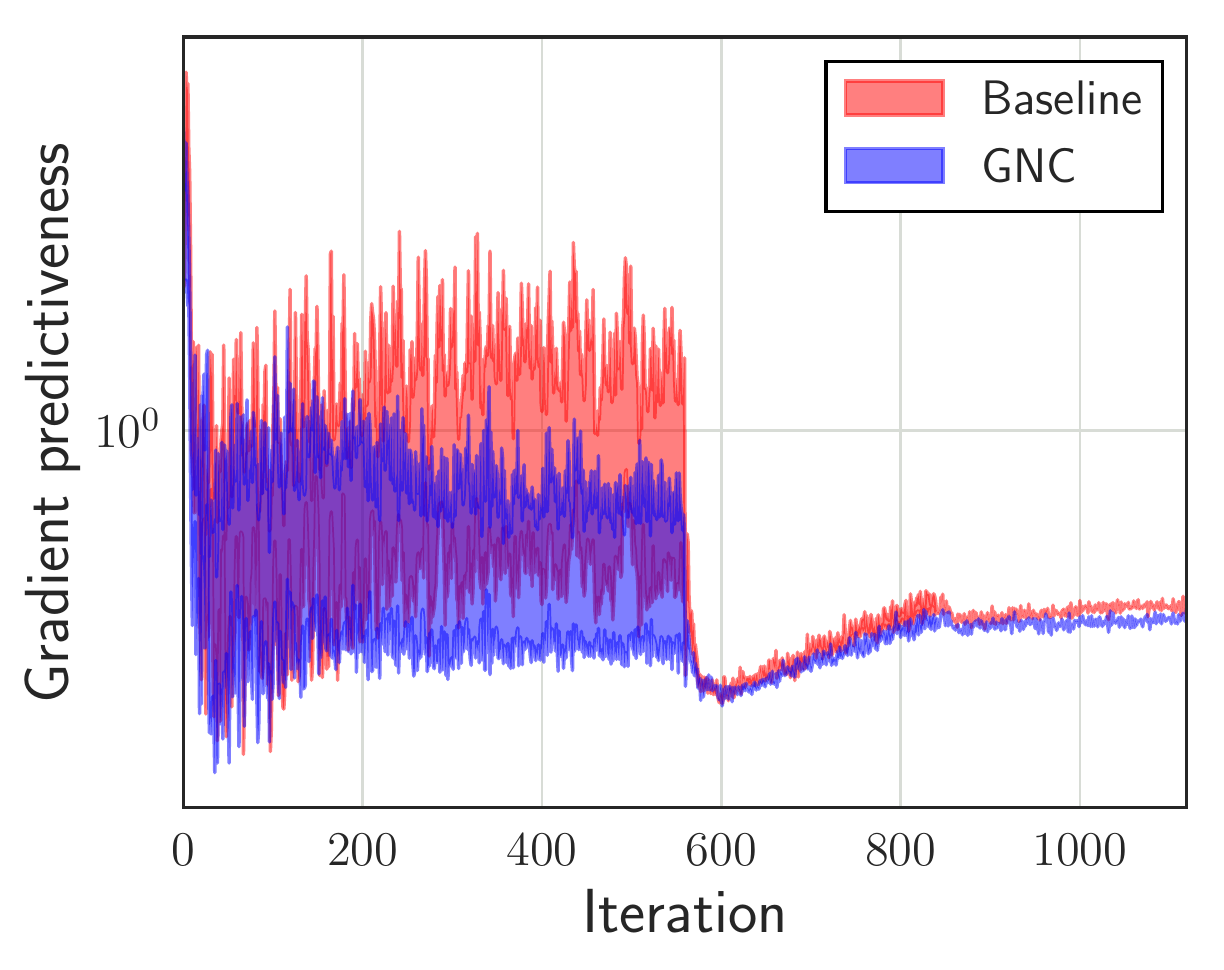}}
    \end{center}
  \end{minipage}
  \begin{minipage}{0.33\hsize}
    \begin{center}
      \def\subfigcapskip{0pt}
      \subfigure[$\beta$ -smoothness]{
        \includegraphics[width=1.8in]{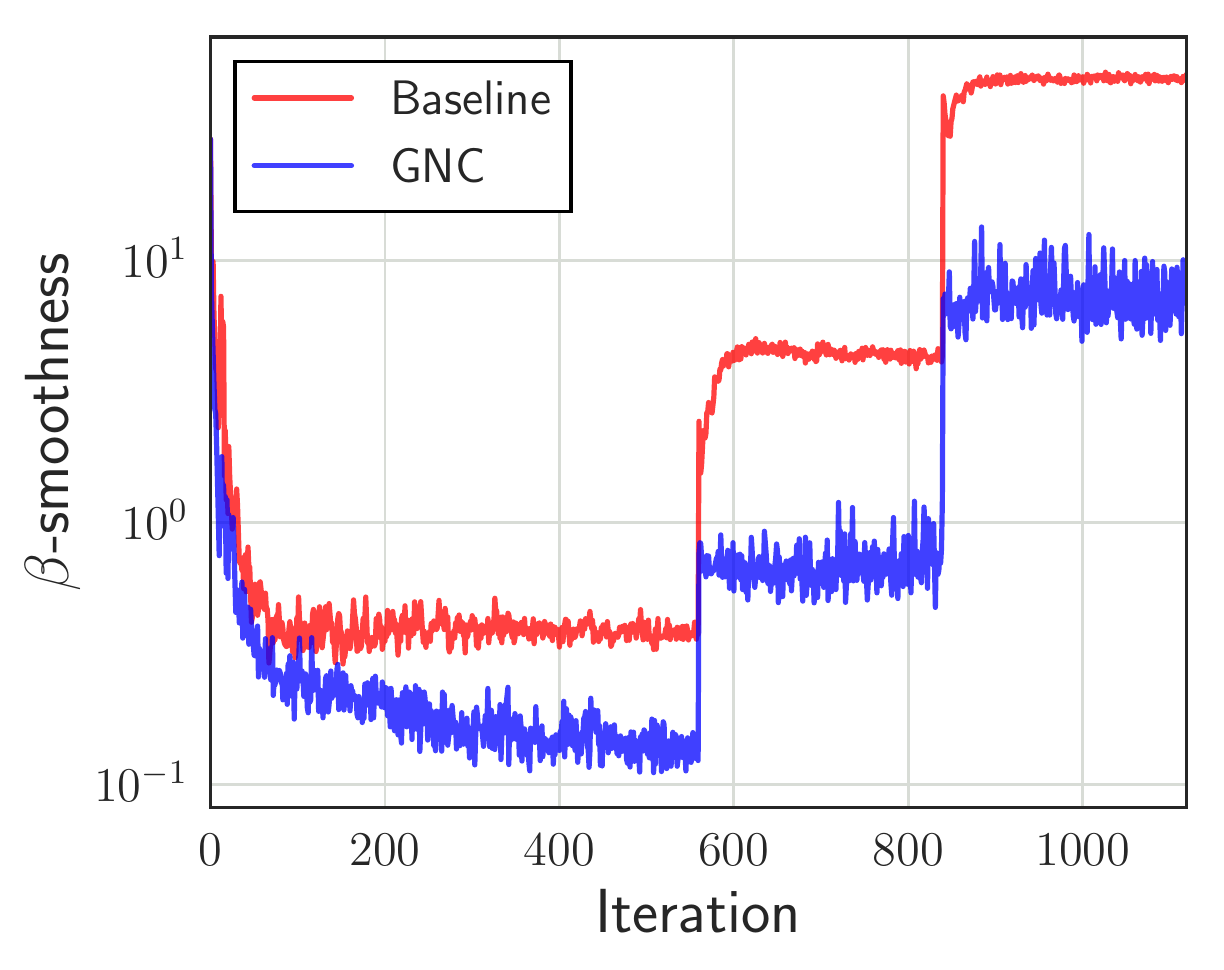}}
    \end{center}
  \end{minipage}
  \vskip -0.15in
  \caption{
    Smoothness of the loss function with and without GNC.
    We calculated variation (shaded regions) in {\bf (a)}~loss, {\bf (b)}~$l_{2}$ changes in the gradient,
    and {\bf (c)}~maximum $l_{2}$ difference in gradient over $l_{2}$ distance.
    These variations and maximum differences were calculated from the range of values along the direction of the gradient.
    Note that in GNC, calculations of loss and gradient along the gradient direction were performed in the GNC method,
    that is, convolved with gradient noise.
    Appendix~\ref{subsec:indicators_for_smoothing_effectiveness} presents detailed calculations of these values.
    All indicators showed clear improvement by GNC.
    Although \citet{Shibani18} showed that BN has a similar effect, GNC provides further improvement.
    GNC seems to have an independent smoothness effect not seen with BN.
  }
  \label{fig:cifar100_lipschizness}
\end{figure}

Fig.~\ref{fig:cifar100_lipschizness}(a) shows variation in loss in the gradient direction.
The shaded region for GNC is narrower than the baseline, indicating that variation in GNC loss is stabilized.
Fig.~\ref{fig:cifar100_lipschizness}(b) shows variation in gradients in the gradient direction.
As in (a), gradient variation is stabilized.
Finally, Fig.~\ref{fig:cifar100_lipschizness}(c) shows gradient stability through the training process.
It is obvious that GNC has a strong smoothing effect for the loss function.
Fig.~\ref{fig:fig_cifar100_b8k_train_loss} shows additional evidence for smoothness by observing the training loss curve.

\subsection{Effectiveness of GNC with Other Optimization Techniques}
\label{subsec:gnc_robustness}

While GNC improves generalization performance by the smoothing effect,
modern optimization techniques for deep learning, such as MM, WD, or DA,
can also improve generalization.
By observing generalization differences between optimization techniques with GNC,
we showed the robustness of GNC for generalization.

\begin{table}[h]
\caption{Validation accuracy improvement by GNC.
We evaluated combinations of GNC with WD, MM, or DA.
In Sec.~\ref{subsec:gnc_robustness}, ``baseline'' indicates no use of GNC, WD, MM, or DA, while
``full'' means full use of the baseline with WD, MM, and DA.
We also evaluated dependency on batch size.
Except for DA, GNC improved validation accuracy.
In the case of DA, we calculated the large-batch gradient with augmented data,
but the full gradient was calculated without augmented data.
This might hinder predicting the full gradient of training data,
eventually dropping generalization performance.
Hence, future work should investigate this mechanism.
}
\label{table_ablation_study_cifar100}
\begin{center}
\begin{small}
\begin{tabular}{cccccccc}
\toprule
Batch Size & GNC & Baseline & +WD & +MM & +DA & Full \\
\midrule
4,096 & off &       42.80$\pm{0.90}$ &       42.63$\pm{1.21}$ &       50.43$\pm{1.21}$ & {\bf 64.57}$\pm{0.19}$ &       72.87$\pm{0.37}$ \\
     & on  & {\bf 46.35}$\pm{0.56}$ & {\bf 47.45}$\pm{0.63}$ & {\bf 51.54}$\pm{1.13}$ &       63.68$\pm{0.15}$ & {\bf 72.89}$\pm{0.38}$ \\ \hline
8,192 & off &       45.66$\pm{1.09}$ &       47.64$\pm{0.45}$ &       45.97$\pm{3.29}$ & {\bf 57.83}$\pm{0.76}$ &       69.21$\pm{2.11}$ \\
     & on  & {\bf 48.74}$\pm{1.06}$ & {\bf 49.73}$\pm{0.81}$ & {\bf 48.67}$\pm{1.01}$ &       56.60$\pm{0.67}$ & {\bf 71.35}$\pm{0.27}$ \\
\bottomrule
\end{tabular}
\end{small}
\end{center}
\end{table}

Table~\ref{table_ablation_study_cifar100} shows accuracy gain by GNC, except for DA.
The accuracy drop in DA might be caused by the augmented data,
which were not included in the calculation of the full gradient.
Accuracy was improved in the case of ``full'', however, even with DA.
There may be interactions among these optimization techniques, so this should be investigated in future work.
As described in Sec.~\ref{subsec:gnc_smoothing_evidence},
we also provide the same evidence for the four combination cases above (Fig.~\ref{fig:cifar100_rubust_lipschizness}).

\section{Performance Evaluations for ImageNet, CIFAR-10, and CIFAR-100}
\label{main_result}

To measure the effect of GNC, we benchmarked against ImageNet-1K~(ImageNet 2012) training,
measuring validation accuracy at the 90th epochs with ResNet50 by a DP-SGD-based method using 1,024 parallel workers.
Some studies have taken on the so-called ``time attack challenge''~(\citet{Goyal17,Iandola15,Akiba17,You18}),
but we focus solely on generalization performance at the target epoch.
Due to the simplicity of GNC, we believe it can be easily merged with these implementational speedup techniques
with no performance drop.

To fairly measure essential performance gains by GNC,
we reproduced current state-of-the-art methods and applied further tuning as the baseline method.
In this case, the baseline method was based on \citet{You18},
adding data augmentation~(\citet{Goyal17}) and our learning rate tuning
(see Appendix~\ref{detail_experimental_setup}).
As a result, our baseline method reached 75.89\% top-1 accuracy,
while \citet{You18} originally reported 75.4\%.
While other researchers have reported higher accuracies (76.2\% in \citet{Jia18}, 76.4\% in \citet{Ying18}),
they do not reveal detailed training settings, placing their results beyond the scope of this paper
(see Table~\ref{app:imagenet_table_benchmark} for details).

Needless to say, the only difference between the baseline method and the GNC (+baseline) method is the use of GNC.
All hyperparameters except those for GNC are the same for both methods.
Furthermore, we eliminated randomness in the training process, as in Sec.~\ref{empirical_analysis}.

\begin{table}[h]
  \caption{Validation accuracies (\%) using various datasets.
  We report the mean and standard deviation of test accuracies among five runs.
  In all cases, GNC or GNCtoRNC improved validation accuracy.
  }
  \label{table_accuracies2}
  \begin{center}
  \begin{small}
  \begin{tabular}{lcccccc}
  \toprule
    Datasets & Batch size & Baseline & RNC & GNC & GNCtoRNC \\
  \midrule
    ImageNet  & 32,768  & 75.89$\pm{0.09}$ & 75.86$\pm{0.13}$ & 76.03$\pm{0.18}$ & {\bf 76.05}$\pm{0.07}$ \\
              & 131,072 & 65.57$\pm{0.45}$ & 65.14$\pm{0.86}$ & {\bf 68.39}$\pm{0.40}$ & 68.13$\pm{0.48}$ \\ \hline
    CIFAR-10  & 4,096   & 93.48$\pm{0.26}$ & 93.70$\pm{0.26}$ & 93.82$\pm{0.64}$ & {\bf 94.00}$\pm{0.60}$ \\
              & 8,192   & 54.51$\pm{7.04}$ & 63.51$\pm{20.45}$& {\bf 91.00}$\pm{1.24}$ & 90.88$\pm{1.53}$ \\ \hline
    CIFAR-100 & 4,096   & 72.87$\pm{0.37}$ & 73.08$\pm{0.27}$ & 72.89$\pm{0.38}$ & {\bf 73.79}$\pm{0.40}$ \\
              & 8,192   & 69.21$\pm{2.11}$ & 70.17$\pm{1.36}$ & 71.35$\pm{0.27}$ & {\bf 71.93}$\pm{0.21}$ \\
  \bottomrule
  \end{tabular}
  \end{small}
  \end{center}
  \vskip -0.1in
  \end{table}
  
Table~\ref{table_accuracies2} shows the results.
In general, GNC improved generalization performance for all datasets.
In ImageNet with a batch size of 32,678,
we achieved a mean score of 76.05\% and best score of 76.29\% with GNC (see Table~\ref{app:table_rawdata_val_acc}),
which is state-of-the-art generalization performance.
From the two observations regarding the latter part of the training process---
the predictiveness drop with the full gradient described in Sec.~\ref{subsec:similarity_fg_lb}
and the high anisotropy of gradient noise described in Fig.~\ref{fig:cifar100_gradient_noise_vs_random_noise}---
gradient noise might be strongly directed in the ``wrong'' direction.
From these observations,
we decided to try switching from GNC to RNC in the latter part of the training process (GNCtoRNC).

GNCtoRNC achieved a more stable performance gain than GNC and improved the ImageNet score.
In the case of 4,096 workers in ImageNet, the performance gain increased even more.
In CIFAR-10 with a batch size of 8,192, the performance gain of both GNC and GNCtoRNC was extremely large.
The reason for this is the instability of the training process by the baseline method.
This also shows the strong smoothing effect of GNC.
Table~\ref{app:table_rawdata_val_acc} gives the detailed analyses for these performance evaluations.

\section{Conclusion}
\label{conclusion}

We proposed a new method called GNC to better generalize large-scale DP-SGD in deep learning.
GNC performs \emph{convolution among parallel workers}, using \emph{gradient noises}
that can be easily computed on each worker
and effectively eliminate sharp local minima.
Experiments demonstrated that using GNC can realize high generalization performance.
For example, GNC achieved 76.29\% peak top-1 accuracy on ImageNet training.
We empirically showed that gradient noises more effectively
smooth the loss function than do isotropic random noises.
We hope that these findings on gradient noise
may contribute to the problem of improving generalization in large-scale deep learning.

\newpage

\subsubsection*{Acknowledgments}

TS was partially supported by MEXT Kakenhi (26280009, 15H05707,
18K19793 and 18H03201), and JSTCREST.
A part of numerical calculations were carried out on the TSUBAME3.0 supercomputer at Tokyo Institute of Technology.

\small
\bibliography{perilla}
\bibliographystyle{plainnat}

\newpage

\normalsize

\appendix

\section{Detailed Experimental Setup}
\label{detail_experimental_setup}

\subsection{Datasets}
\label{subsec:dataset}

We empirically evaluated the performance of GNC in large scale distributed training
with standard image classification datasets,
performing experiments with CIFAR-10, CIFAR-100~(\citet{CIFAR09}),
and ImageNet-1K (ImageNet 2012, \citet{ImageNet15}).
In ImageNet-1K, consisting of 1000 classes,
we trained the model on 1.28 million training images,
and evaluated top-1 validation accuracies on 50,000 validation images.
We also evaluated with CIFAR-10 and CIFAR-100, consisting of 10 and 100 classes, respectively.
The model was trained on 50,000 training images,
and evaluated on 10,000 validation images.

\begin{table}[h]
\caption{Basic settings for large-scale distributed training. We denote the batch size per worker as the small-batch size.}
\label{table_basical_settings}
\begin{center}
\begin{small}
\begin{tabular}{lccccc}
\toprule
	Dataset & Network & Epochs & Batch size & Small-batch size & \# of workers \\
\midrule
	ImageNet & ResNet-50 & 90 & 32,768 & 32 & 1,024 \\ \cline{4-6}
		 &           &    & 131,072 & 32 & 4,096 \\ \hline
	CIFAR-10/CIFAR-100 & ResNet-32 & 160 & 4,096 & 32 & 128 \\ \cline{4-6}
			   &           &     & 8,192 & 32 & 256 \\
\bottomrule
\end{tabular}
\end{small}
\end{center}
\vskip -0.1in
\end{table}

\begin{table}[h]
\caption{Common settings for learning rate control. We adopted a gradual warmup, in which the learning rate
	gradually ramped up from ``GW start lr'' to ``Initial lr'' over the first ``GW epochs'' in the
	training.}
\label{table_lr_settings}
\begin{center}
\begin{small}
\begin{tabular}{lcccccc}
\toprule
	Dataset & LARS & Decay rule & GW epochs & GW start lr & Batch size & Initial lr \\
\midrule
	ImageNet & $\surd$ & Polynomial & 5 & 1.0 &  32,768 & 23.0  \\ \cline{6-7}
		 &         &            &   &     & 131,072 & 42.0  \\ \hline
	CIFAR-10/CIFAR-100 & $\times$ & Step & 10 & 0.025 & 4,096 & 3.2  \\ \cline{6-7}
			   &          &      &    &       & 8,192 & 6.4 \\
\bottomrule
\end{tabular}
\end{small}
\end{center}
\vskip -0.1in
\end{table}

\begin{table}[h]
\caption{ $\alpha$ settings for each methods with noise.}
\label{table_alpha_settings}
\begin{center}
\begin{small}
\begin{tabular}{lcccc}
\toprule
	Dataset & Batch size & RNC & GNC & GNCtoRNC \\
\midrule
	ImageNet  &  32,768 & 0.01  & 0.0001 & GNC:0.01, RNC:0.01 \\
		  & 131,072 & 0.01  & 0.0001 & GNC:0.01, RNC:0.01 \\ \hline
	CIFAR-10  &   4,096 &  0.01 & 0.1    & GNC:0.1, RNC:4.0 \\
		  &   8,192 &  0.01 & 0.1    & GNC:0.1, RNC:2.0 \\ \hline
	CIFAR-100 &   4,096 & 0.001 & 0.1    & GNC:0.1, RNC:2.0 \\
		  &   8,192 &  0.01 & 0.1    & GNC:0.1, RNC:4.0 \\
\bottomrule
\end{tabular}
\end{small}
\end{center}
\vskip -0.1in
\end{table}

\subsection{Noise Scaling}
\label{subsec:noise_scaling}

In general, the norm of the weights differs for each filter in deep learning models.
When giving noise to weights for the purpose of observing the change of the loss,
some filters can be catastrophically affected with constant scale of noise.
To address this problem, \citet{Li17} proposed filter-wise noise scaling and achieved appropriate visualization.
Beyond this intuition,
\citet{Wen18} proposed filter-wise scaling for injecting uniform random noise to the loss function,
which was called \emph{AdaSmoothOut}.
For RNC, we also applied filter-wise noise scaling.
Considering this scaling for a filter $p$,
\Eqref{eq:gradient_convolved_function_r2} with $\alpha$ is deformed as
    \begin{equation}
      \nabla \tilde{f}(d_t;x_{t}^{(p)}) = 
        \frac{1}{M} \sum_{i=1}^{M} \nabla f(d_{t}^{i};x_{t}^{(p)} - \alpha {\eta}_t \|x_{t}^{(p)}\| \frac{{\omega}_{t}^{i(p)}}{\|{\omega}_{t}^{i(p)}\|}),
    \label{eq:rnc_filter_wise_scaling}
    \end{equation}
where $x_{t}^{(p)}$ denotes the weights of filter $p$ and
${\omega}_{t}^{i(p)}$ denotes noise injected into $x_{t}^{(p)}$.
In GNC, in order to clarify the difference in the effect
stemming from random noise and gradient noise,
filter-wise scaling was introduced as
    \begin{equation}
      \nabla \tilde{f}(d_t;x_{t}^{(p)}) = 
        \frac{1}{M} \sum_{i=1}^{M} \nabla f(d_{t}^{i};x_{t}^{(p)} - \alpha {\eta}_t \|x_{t}^{(p)}\| {\omega}_{t}^{i(p)}).
    \label{eq:gnc_filter_wise_scaling}
    \end{equation}

\subsection{Experimental Setup}

We evaluated and compared validation accuracies for GNC and RNC (Sec.~\ref{subsec:RNC})
using conventional DP-SGD with zero noise as a baseline.
In our experiments, we adopted the {\bf noise scaling} version described in Sec.~\ref{subsec:noise_scaling} for
GNC and RNC.

For all methods, we adopted standard SGD with momentum.
There are various momentum implementations, as mentioned in \citet{Goyal17},
but we used the form
    \begin{eqnarray*}
      v_t &=& m v_{t - 1} - {\eta}_t \nabla f(d_t;x_{t}) \\
      x_{t+1} &=& x_{t} + v_t
    \label{eq:momentum_update}
    \end{eqnarray*}
where $v_t$ denotes the update vector,
${\eta}_t$ denotes the learning rate, and $m$ denotes the momentum decay factor.
We used a momentum decay factor $m$ of $0.9$.
Experiments were performed under the large-scale distributed settings shown in Table~\ref{table_basical_settings}.
In following sections, we describe the detailed settings for each dataset.

\begin{figure}[h]
  \begin{minipage}{0.5\hsize}
    \begin{center}
      \subfigure[LRS for CIFAR-10/CIFAR-100]{
        \includegraphics[width=2.5in]{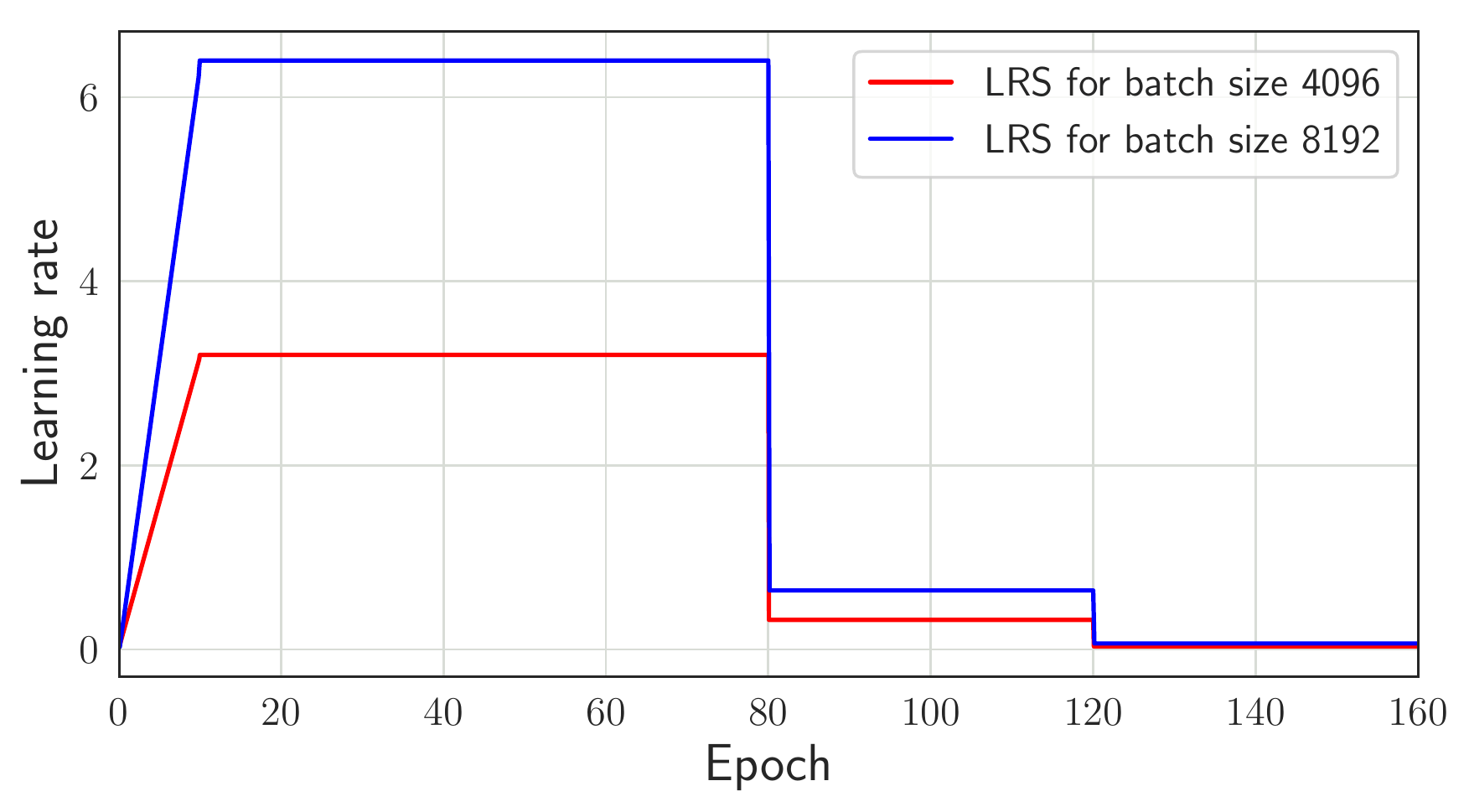}}
    \end{center}
  \end{minipage}
  \begin{minipage}{0.5\hsize}
    \begin{center}
      \subfigure[LRS for ImageNet]{
        \includegraphics[width=2.5in]{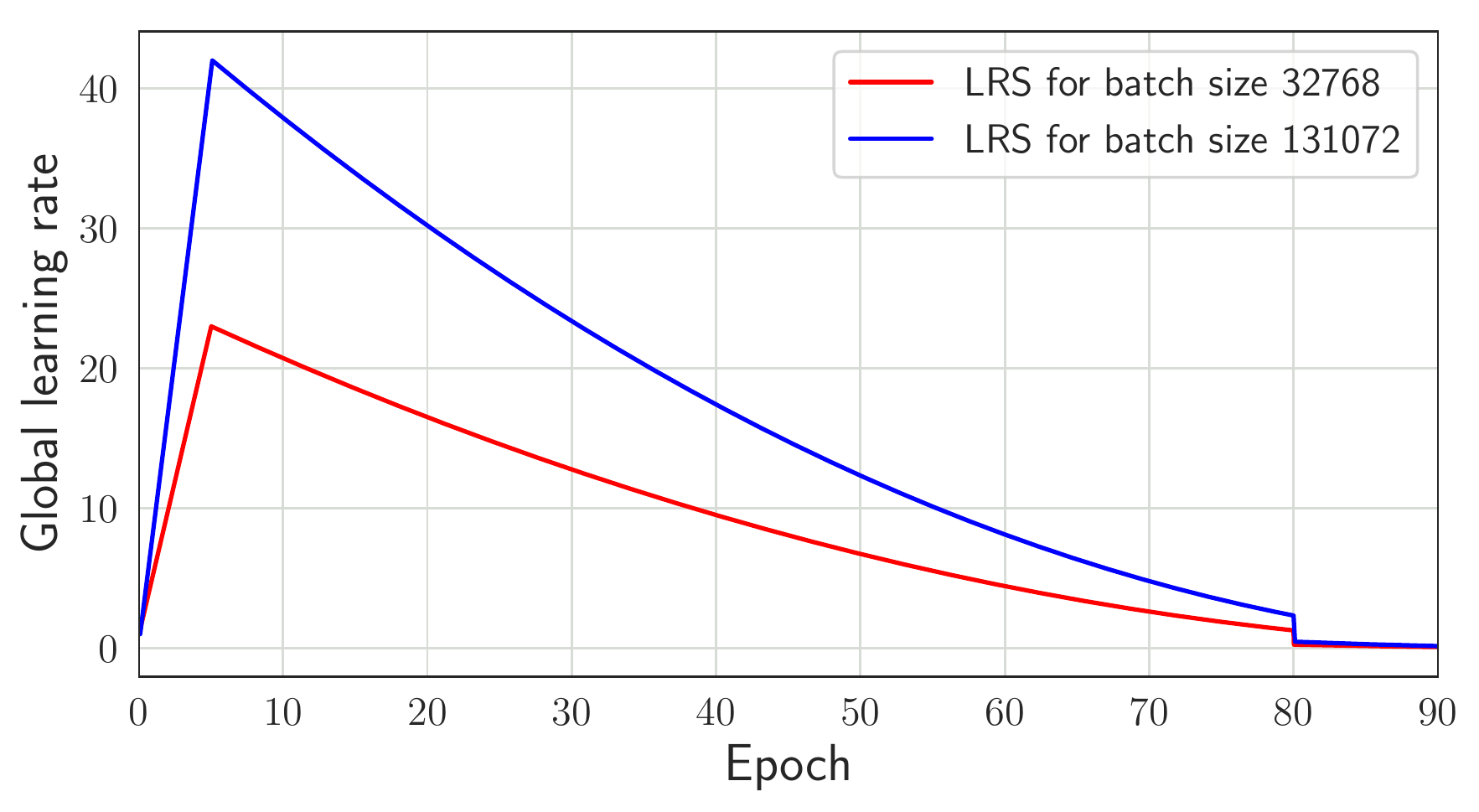}}
    \end{center}
  \end{minipage}
	\caption{(a)~Learning rate scheduling for CIFAR-10/CIFAR100. We adopted gradual warmup over the
	first ten epochs, after which the learning rate was divided by 10 at the 80th and 120th epochs.
	(b)~Scheduling of global learning rate ${\gamma}_t$ for ImageNet.
        We adopted a gradual warmup over the first five epochs, after which ${\gamma}_t$ was scheduled
        based on the polynomial decay policy and divided by 5 at the 80th epochs.
        }
  \label{fig:learning_rate_scheduling}
\end{figure}


\subsection{CIFAR-10/CIFAR-100}
\label{subsec:cifar10_100_detail}
We basically followed the experimental settings reported in~\citet{He15}.
Namely, the model was trained for
160 epochs, with the learning rate divided by 10 only at the 80th and 120th epochs.
To restrain early-stage instability in large-batch training,
we adopted a gradual warmup.
See Table~\ref{table_lr_settings} and Fig.~\ref{fig:learning_rate_scheduling} (a) for details.
In practice, these GW start lr and initial lr follow the linear scaling rule (\citet{Goyal17}),
based on a value of $0.1$ at a batch size of 128 (\citet{He15}).
As an exception, we adopted GW start lr of $0.00025$ for the experiment in
Fig.~\ref{fig:cifar100_full_gradient_similarity} (c), because the experiment was executed without BN.
We thus could not use a larger learning rate, and so used corresponding initial lr of 0.064, 0.032, and 0.016 for
batch sizes 8K, 4K, and 2K, respectively.
We used a weight decay of $0.0001$.
Following the standard data augmentation practices described in \citet{He15},
we used horizontal flip and random crop.
We also used {\it random erasing} with the hyperparameters proposed by \citet{Zhong17}.
We tuned $\alpha$ in 
\Eqref{eq:gnc_filter_wise_scaling} and \Eqref{eq:rnc_filter_wise_scaling} for each method with noise and 
for each batch size, respectively. See Table~\ref{table_alpha_settings} for details.
For the GNCtoRNC method, we started with GNC and swithed to RNC at the 120th epochs.

\subsection{ImageNet}
\label{subsec:imagenet_detail}

The model was trained for 90 epochs. We used layer-wise adaptive rate scaling (LARS)
proposed by~\citet{You17}.
The LARS algorithm computes the learning rate for each layer.
The learning rate $\eta_{t}^{(l)}$,
which is called \emph{local learning rate} in \citet{You17}, used for layer $l$ is
    \[
      {\eta}_{t}^{(l)} = 
        {\gamma}_t
        \tau
        \frac{\|x_{t}^{(l)}\|}{\|\nabla f(d_t;x_{t}^{(l)})\| + \lambda \|x_{t}^{(l)}\|}
    \label{eq:lars}
    \]
where $x_{t}^{(l)}$ denotes the weights for layer $l$,
${\gamma}_t$ a \emph{global learning rate},
$\tau$ a fixed LARS coefficient typically set to $0.001$,
and $\lambda$ a weight decay coefficient.
We used a weight decay $\lambda$ of $0.0001$.
For the scheduling policy of the global learning rate ${\gamma}_t$,
we used polynomial decay as ${\gamma}_{t} = {\gamma}_{\rm initial} (1 - \frac{t}{T})^2$ with
$T = 4,120$. 
As in the case of CIFAR, we used a gradual warmup with the settings shown in Table~\ref{table_lr_settings}.
Additionally, the global learning rate was divided by 5 after the 80th epochs,
proposed by \citet{Condreanu17} as \emph{final collapse}
(see Fig.~\ref{fig:learning_rate_scheduling} (b) for details).
For training, we followed the techniques for data augmentation used in~\citet{Goyal17}.
We found an appropriate $\alpha$ in \Eqref{eq:gnc_filter_wise_scaling}
and \Eqref{eq:rnc_filter_wise_scaling} to achieve better validation accuracies
(see Table~\ref{table_alpha_settings} for details).
For the GNCtoRNC method, we started with GNC and switched to RNC at the 80th epochs.

\subsection{Indicators for Effectiveness of Loss Function Smoothing}
\label{subsec:indicators_for_smoothing_effectiveness}

In Sec.~\ref{subsec:gnc_smoothing_evidence}, following \citet{Shibani18},
we introduced three indicators: {\it loss stability}, {\it gradient predictiveness}
and the {\it Lipschitzness of the gradient} to evaluate the effectiveness of loss function smoothing.

During training, we took additional steps in the direction of the gradient at each iteration.
We chose eight steps of a length are evenly taken
from the range [1/2, 2] $\times$ {\it lerning rate} at the iteration.
We calculate the loss values and gradients at each step.
Using these loss values and gradients, {\it loss stability} is expressed by 
the range of the loss values, and {\it gradient predictiveness} is expressed by the range of
Euclidean distances from the original gradients to the gradients at an additional step.
{\it Lipschitzness of the gradient} is expressed by
the maximum value of $L$ satisfying $|\nabla f(x_0) - \nabla f(x_s)| \leq L\| x_0 - x_s \|$, for all $x_s$,
in which $x_0$ denotes the parameter with no additional steps.

\section{Experimental Environment}
\label{sec:experimental_environment}

We implemented our algorithm and all test cases within an existing SW framework,
called Chainer (\citet{Tokui15}) and ChainerMN (\citet{Fukuda17}).
We executed ImageNet training with 64 Tesla V100 GPUs on TSUBAME3.0 (\url{https://www.gsic.titech.ac.jp/en/tsubame}).
We simulated 1,024 or 4,096 parallel workers using 64 GPUs for the ImageNet cases.
In CIFAR-10 and -100 cases, execution used 8 Tesla V100 GPUs on RAIDEN (\url{https://aip.riken.jp/pressrelease/raiden180420/?lang=en}).
We simulated 128 or 256 parallel workers with 8 GPUs for the CIFAR cases.

\section{Additional Results}
\label{sec:additional_results}

\subsection{Ommited Figures}
\label{subsec:ommited_figures}

Additional figures for the analysis in Sec.~\ref{empirical_analysis} are presented below.

\begin{figure}[h]
  \begin{minipage}{0.5\hsize}
    \begin{center}
      \subfigure[Validation accuracy]{
        \includegraphics[width=2.5in]{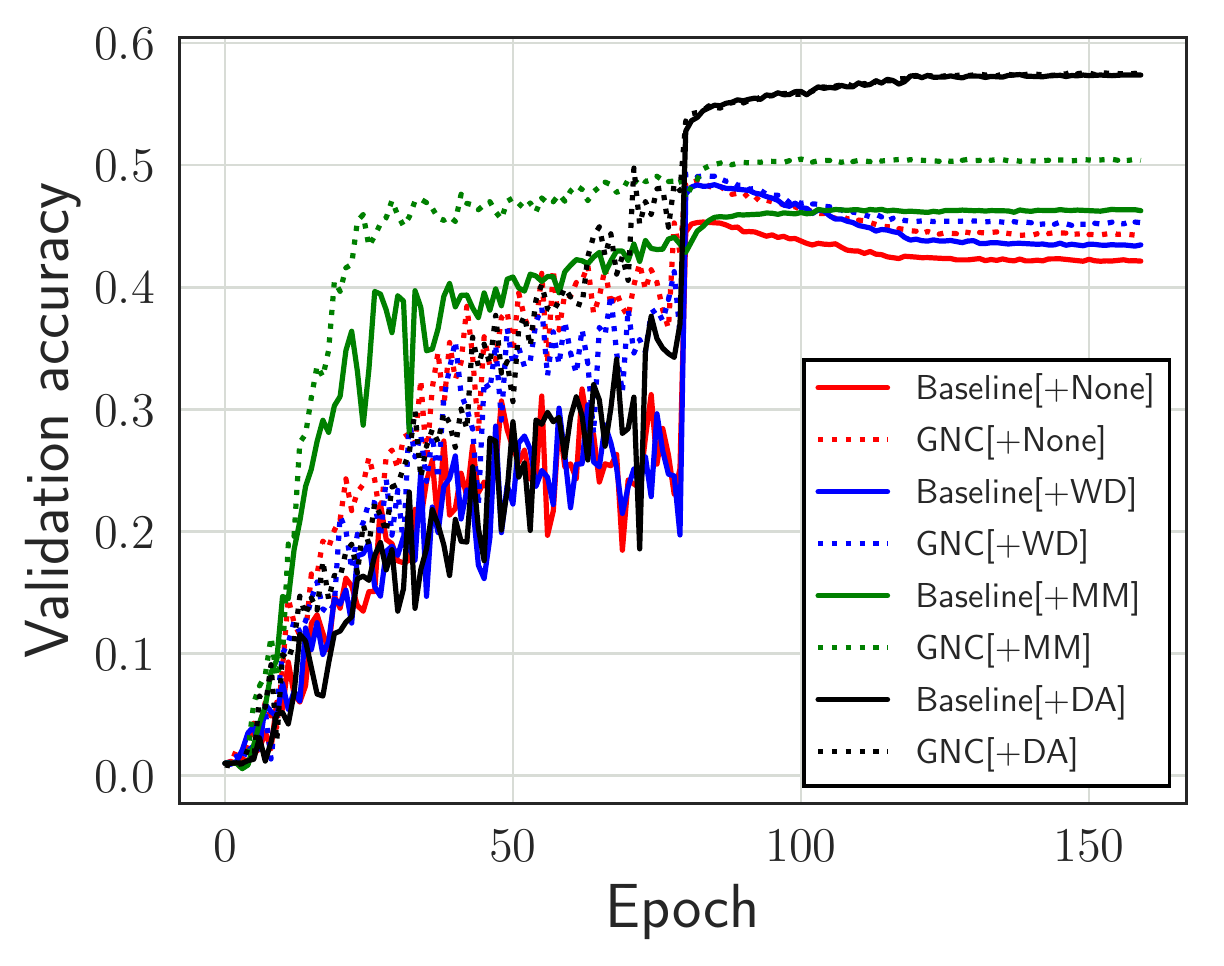}}
    \end{center}
  \end{minipage}
  \begin{minipage}{0.5\hsize}
    \begin{center}
      \subfigure[FG similarity]{
        \includegraphics[width=2.5in]{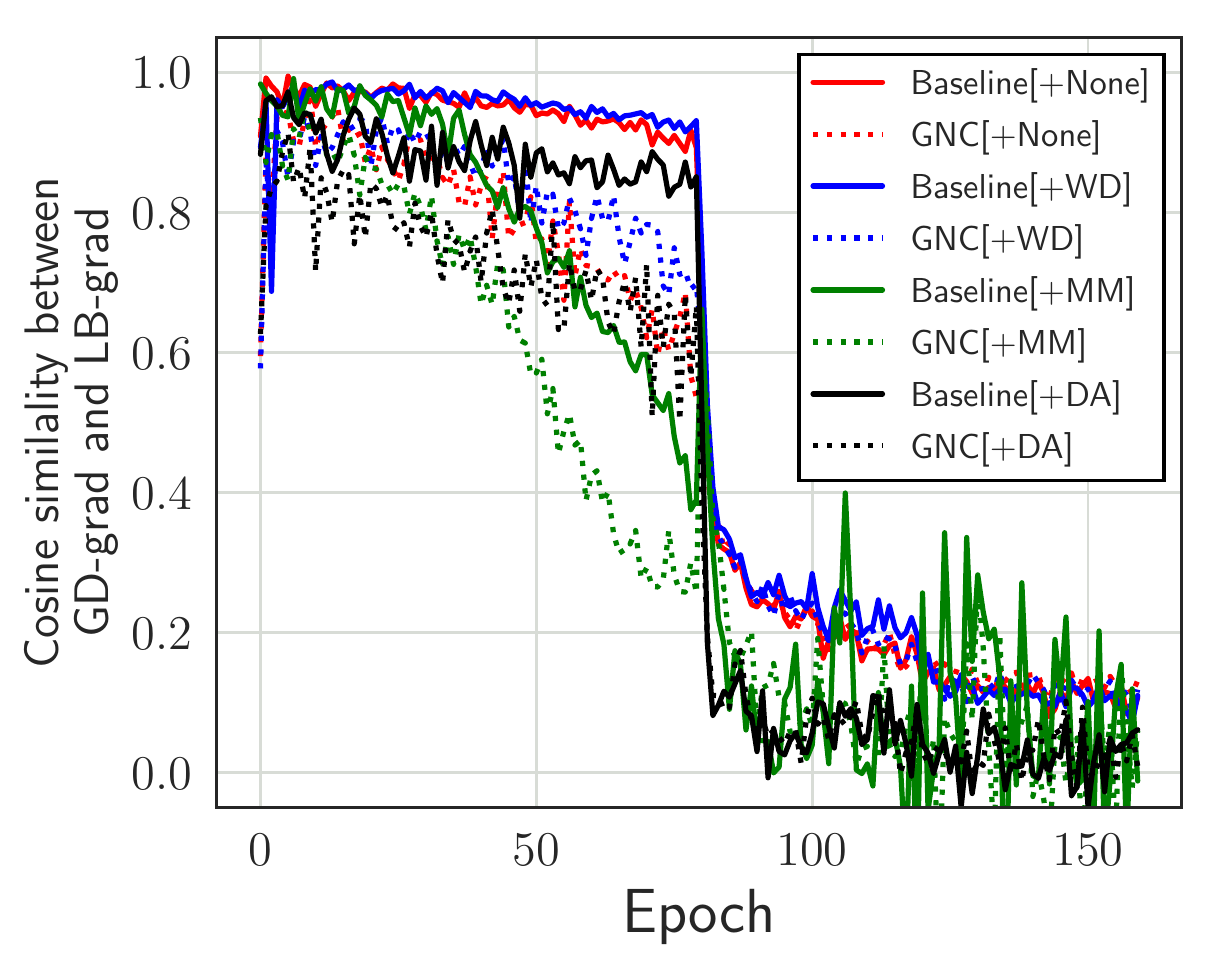}}
    \end{center}
  \end{minipage}
	\caption{
    {\bf (a)}~Validation accuracies of training with and without GNC for the varieties
    in Sec.~\ref{subsec:gnc_robustness};
    results for CIFAR-100 with batch size 8,192.
    {\bf (b)}~Cosine similarity between full gradient (FG)
    and large-batch gradient as calculated by the models in (a).
    Predictiveness of the full gradient dropped at step decay in all test cases.
    }
  \label{fig:robust_valacc_cossim}
\end{figure}

\begin{figure}[h]
  \begin{minipage}{0.5\hsize}
    \begin{center}
      \subfigure[ImageNet losses in 21 epochs]{
        \includegraphics[width=2.5in]{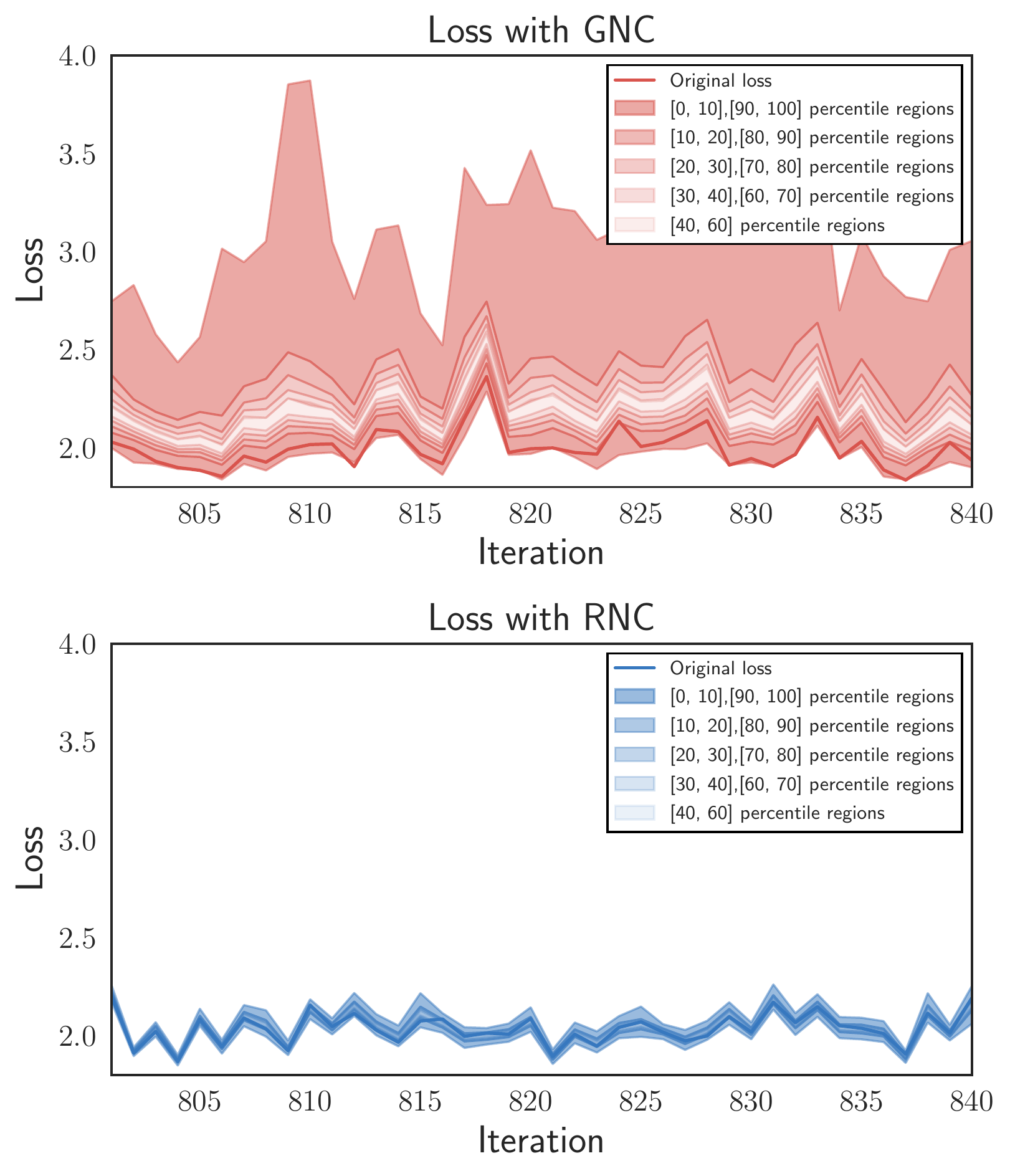}}
    \end{center}
  \end{minipage}
  \begin{minipage}{0.5\hsize}
    \begin{center}
      \subfigure[ImageNet losses in 81 epochs]{
        \includegraphics[width=2.5in]{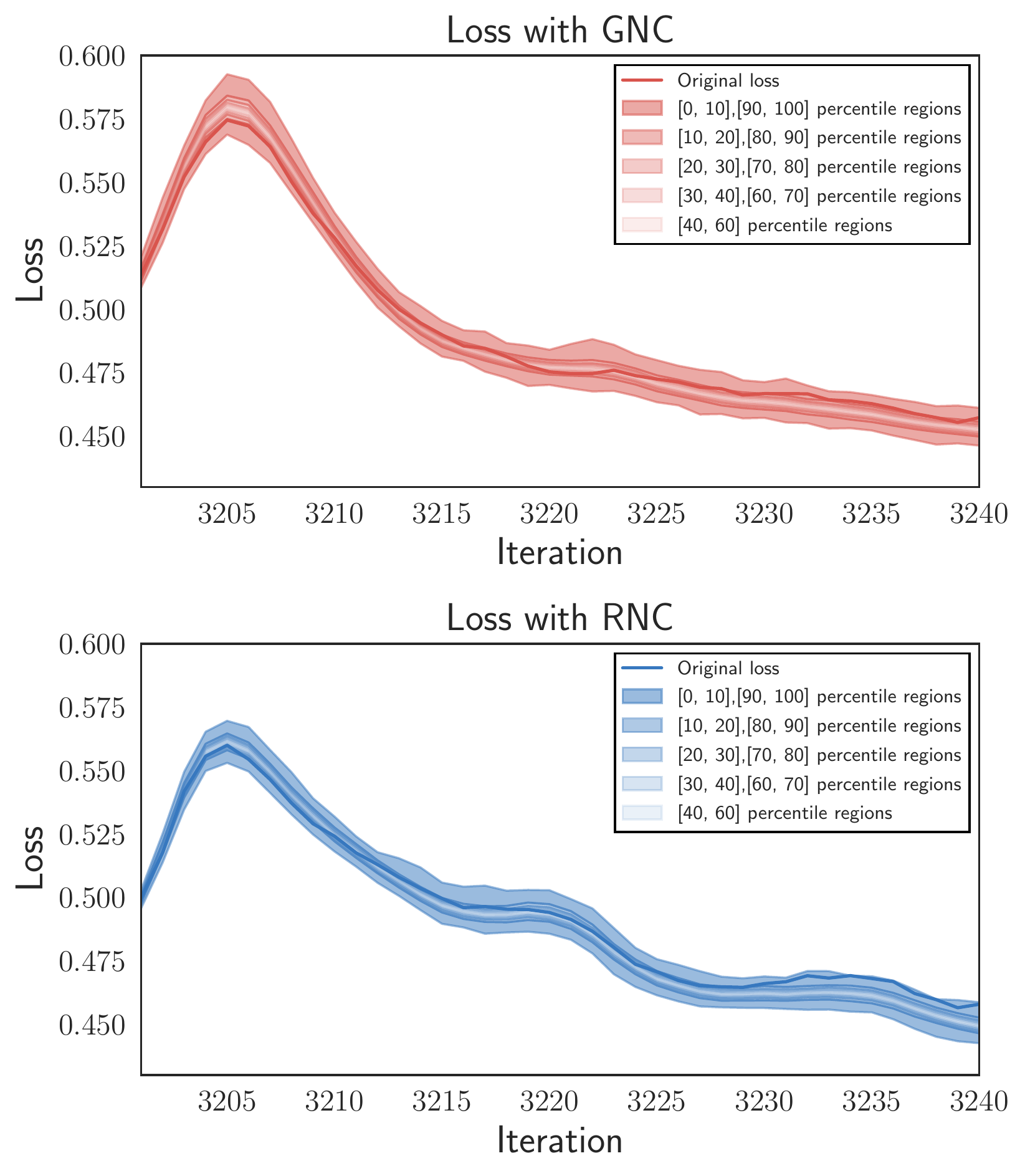}}
    \end{center}
  \end{minipage}
	\caption{
		We measured losses of models $x_{t}^{i}$ in Fig.~\ref{fig:convolution_among_workers_and_iterations}
    for ImageNet training with a batch size of 32,768.
    Loss calculations are the same as in Fig.~\ref{fig:cifar100_gradient_noise_vs_random_noise}.
    The bold line indicates loss of $x_{t}$ through iteration $t$.
    The shaded region indicates losses of $x_{t}^{i}$ among workers $i (i=1,...,M)$.
    If noise ${\omega}_{t}^{i}$ is correctly oriented toward the sharper direction,
    losses $x_{t}^{i} (i=1,...,M)$ vary from the loss of $x_{t}$.
    {\bf (a):}~Losses around the 21the epochs for GNC (upper) and RNC (lower).
		GNC noise can correctly detect sharper directions, so the losses are widely varying.
    In contrast, because RNC noise is random and isotropic, losses do not vary to the extent of GNC.
    {\bf (b):}~Losses around the 81th epochs. GNC cannot detect sharper directions, so
		losses do not vary much for either GNC or RNC.
  }
  \label{fig:imagenet_loss_distribution}
\end{figure}

\begin{figure}[h]
  \begin{minipage}{0.33\hsize}
    \begin{center}
      \subfigure[Baseline]{
        \def\subfigcapskip{0pt}
        \includegraphics[width=1.8in]{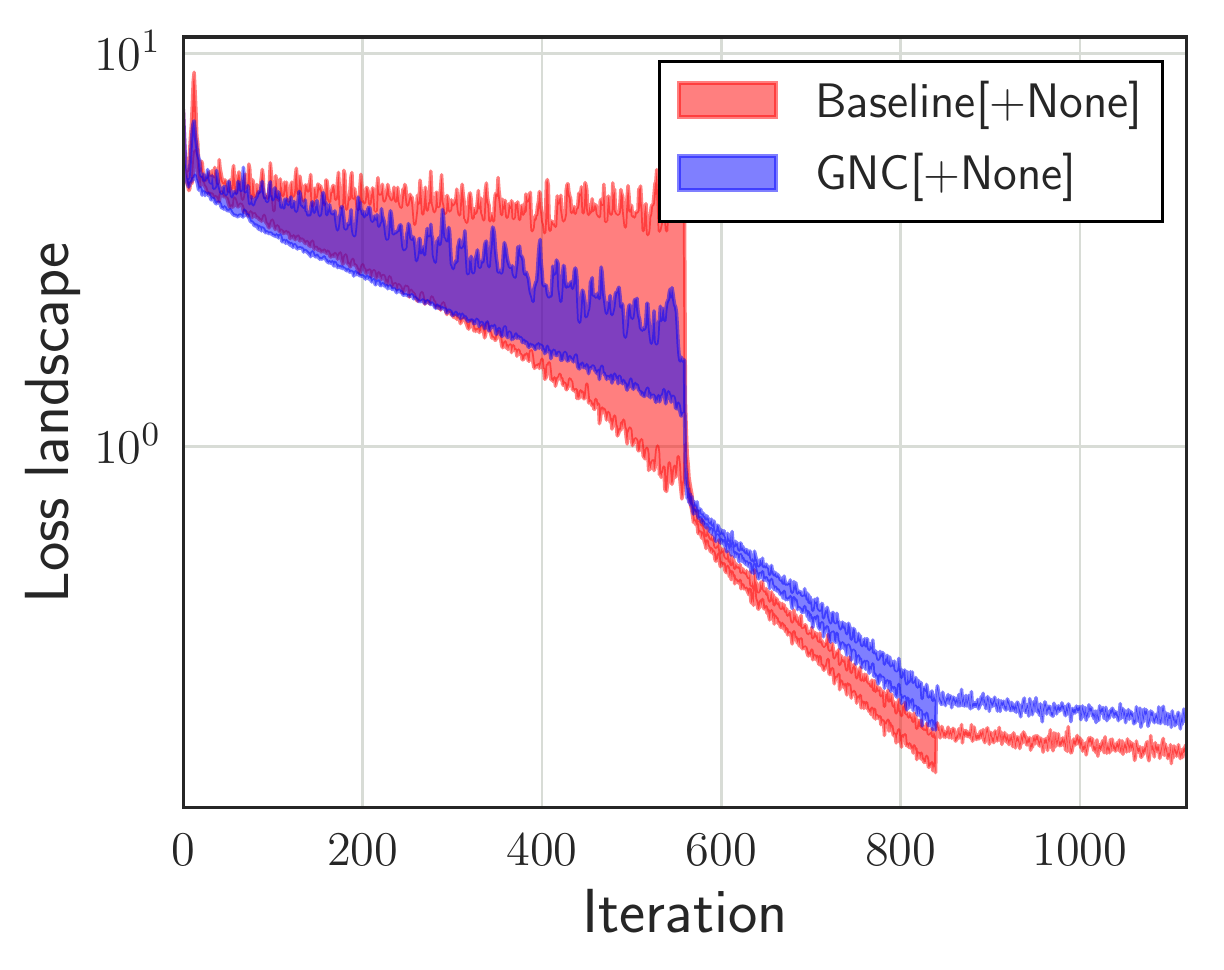}}
      \subfigure[Baseline + WD]{
        \def\subfigcapskip{0pt}
        \includegraphics[width=1.8in]{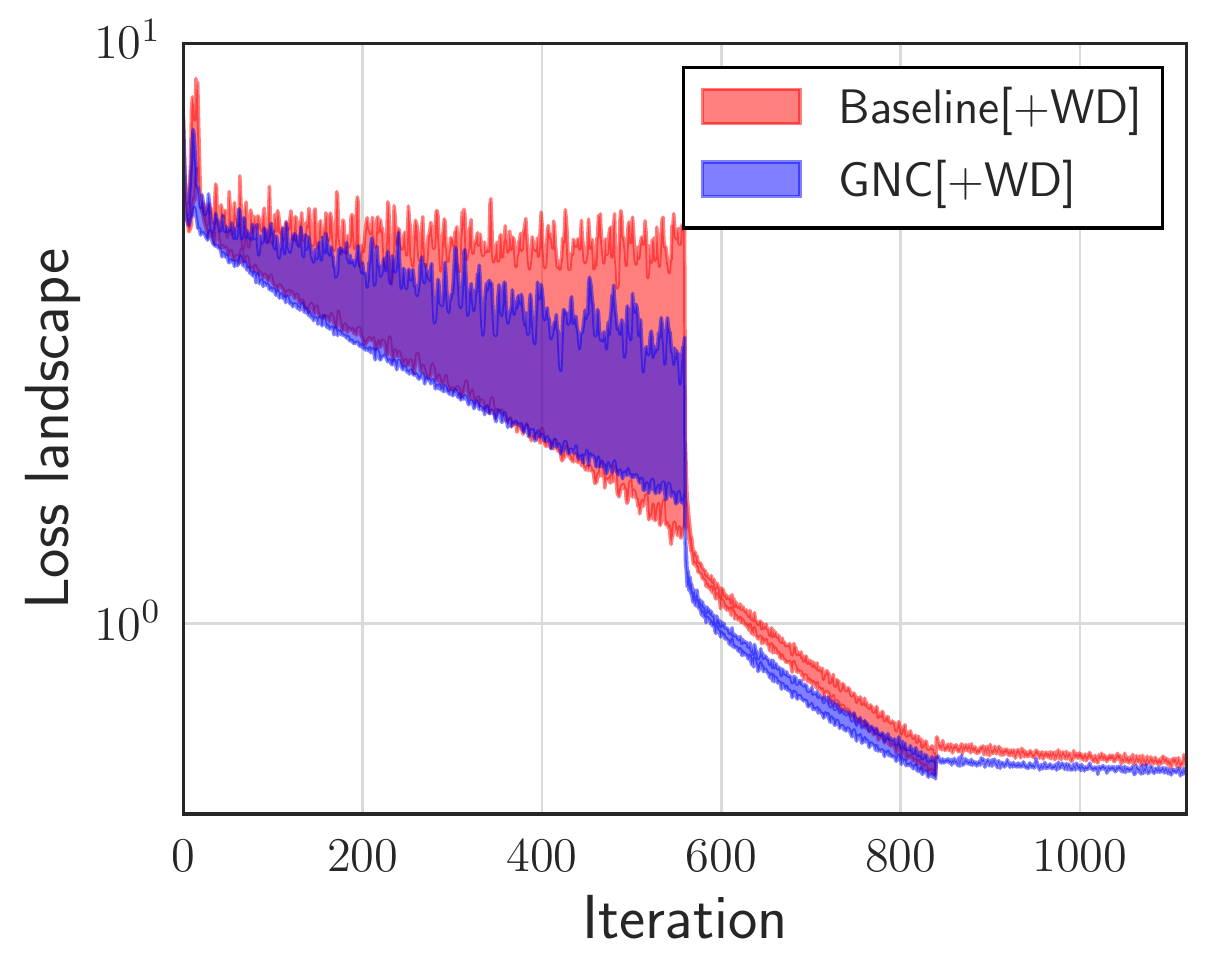}}
      \subfigure[Baseline + MM]{
        \def\subfigcapskip{0pt}
        \includegraphics[width=1.8in]{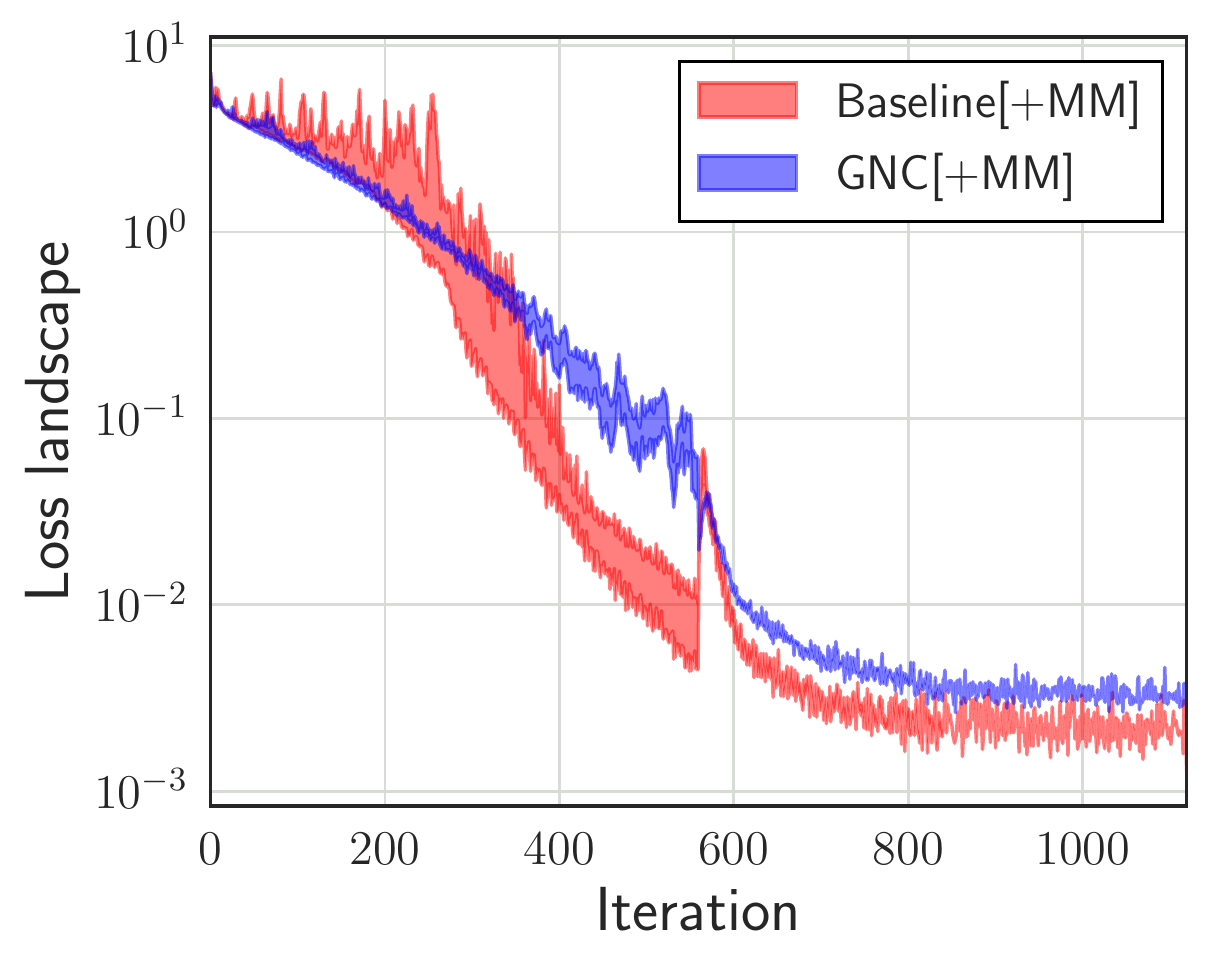}}
      \subfigure[Baseline + DA]{
        \def\subfigcapskip{0pt}
        \includegraphics[width=1.8in]{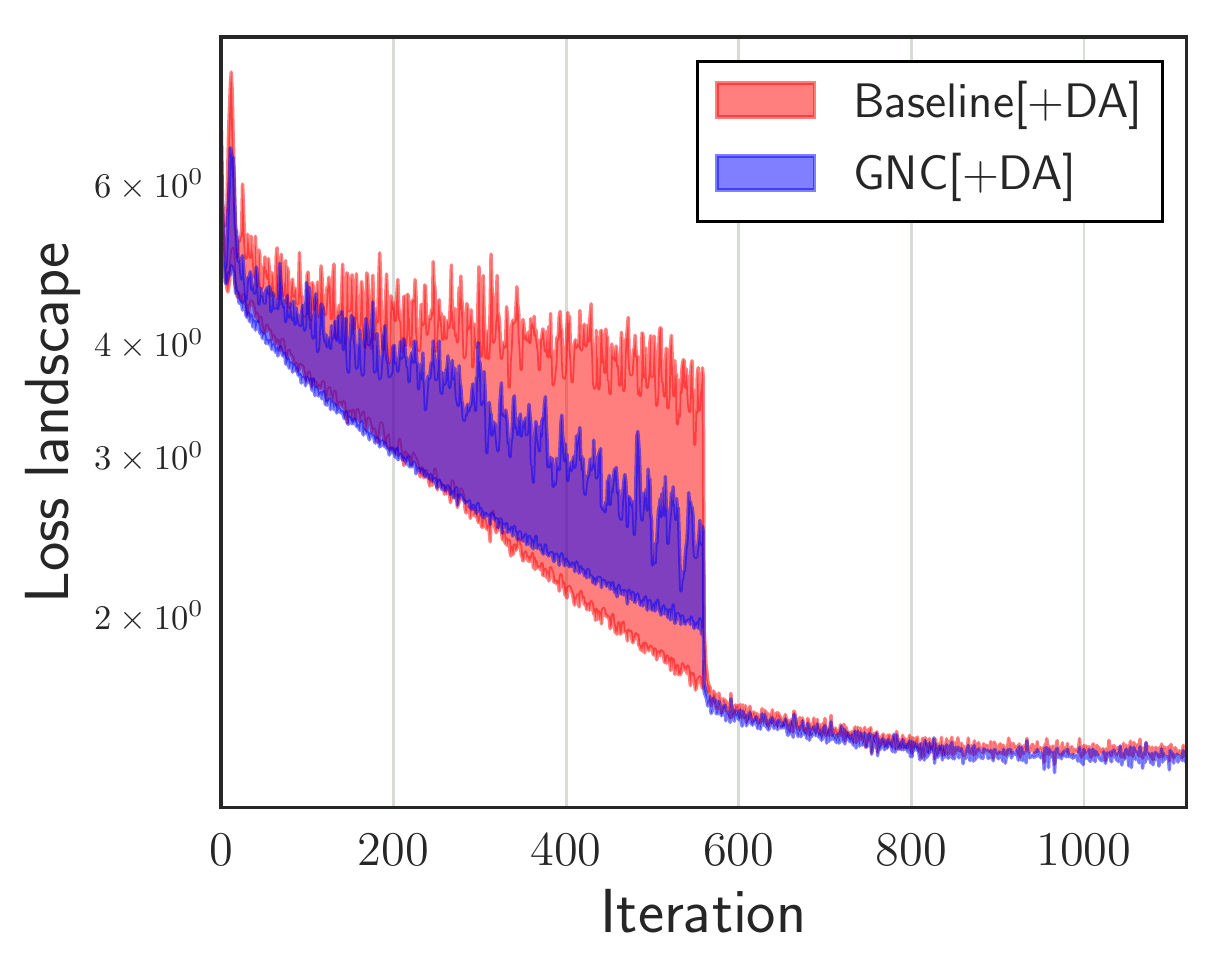}}
    \end{center}
  \end{minipage}
  \begin{minipage}{0.33\hsize}
    \begin{center}
      \subfigure[Baseline]{
        \def\subfigcapskip{0pt}
        \includegraphics[width=1.8in]{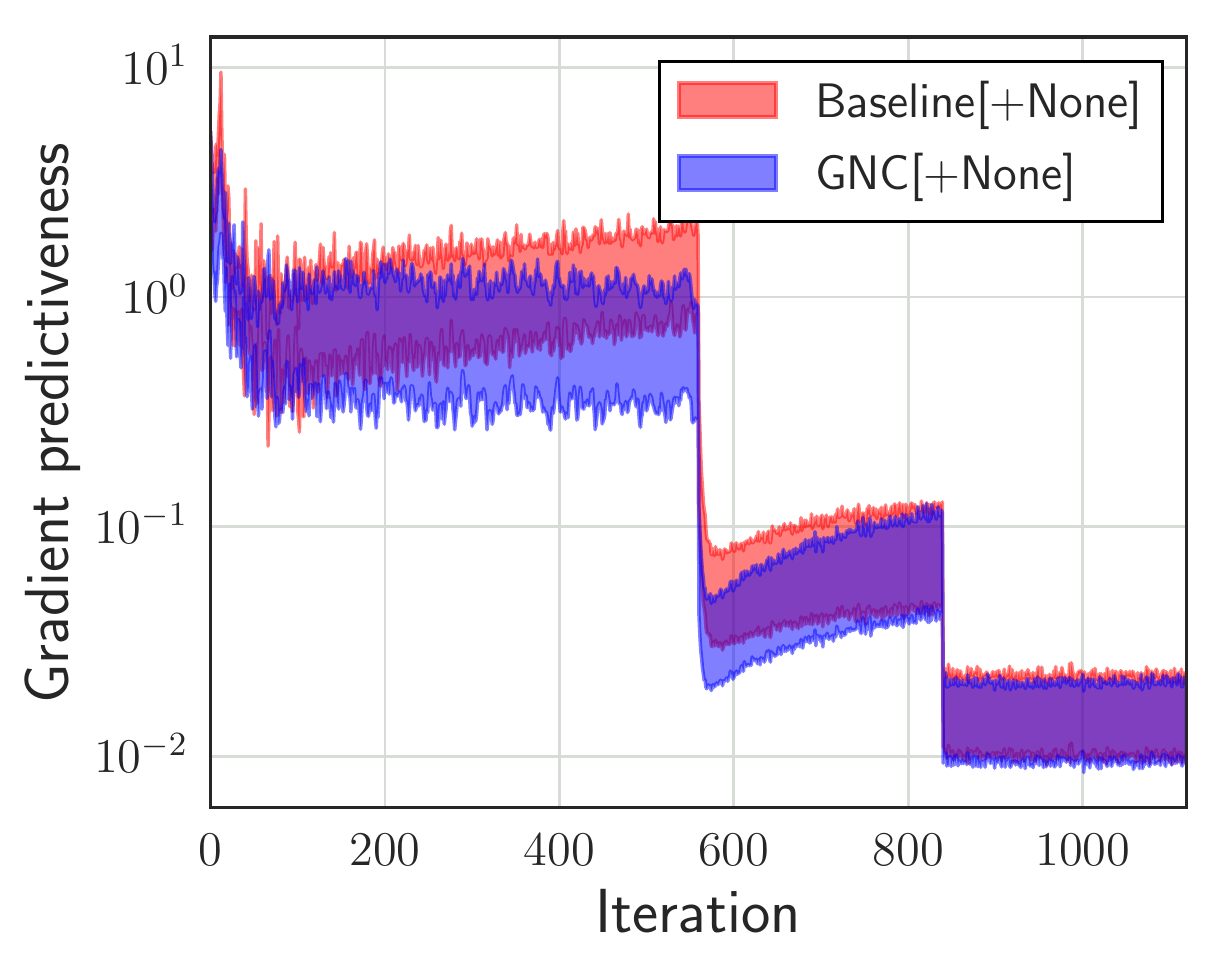}}
      \subfigure[Baseline + WD]{
        \def\subfigcapskip{0pt}
        \includegraphics[width=1.8in]{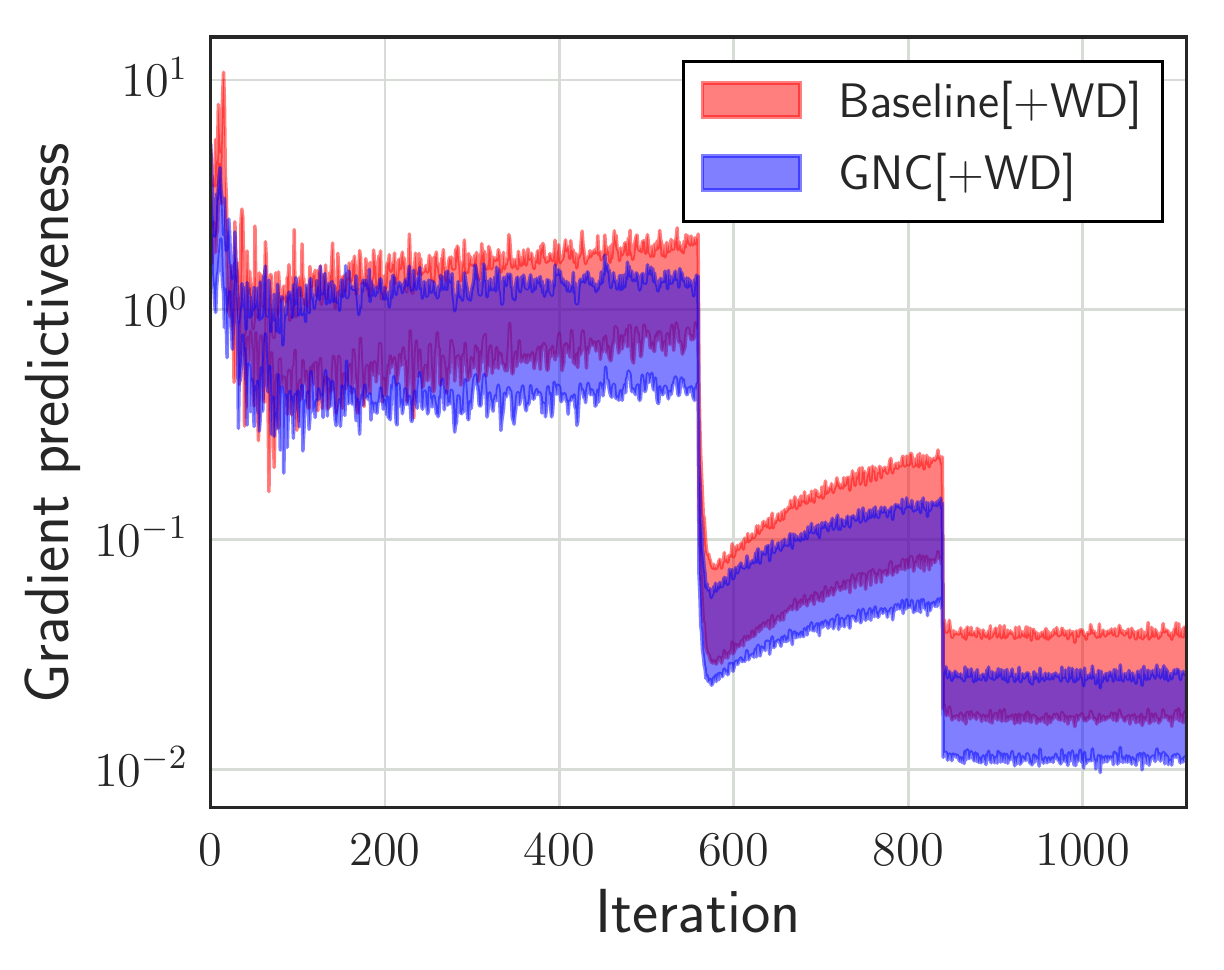}}
      \subfigure[Baseline + MM]{
        \def\subfigcapskip{0pt}
        \includegraphics[width=1.8in]{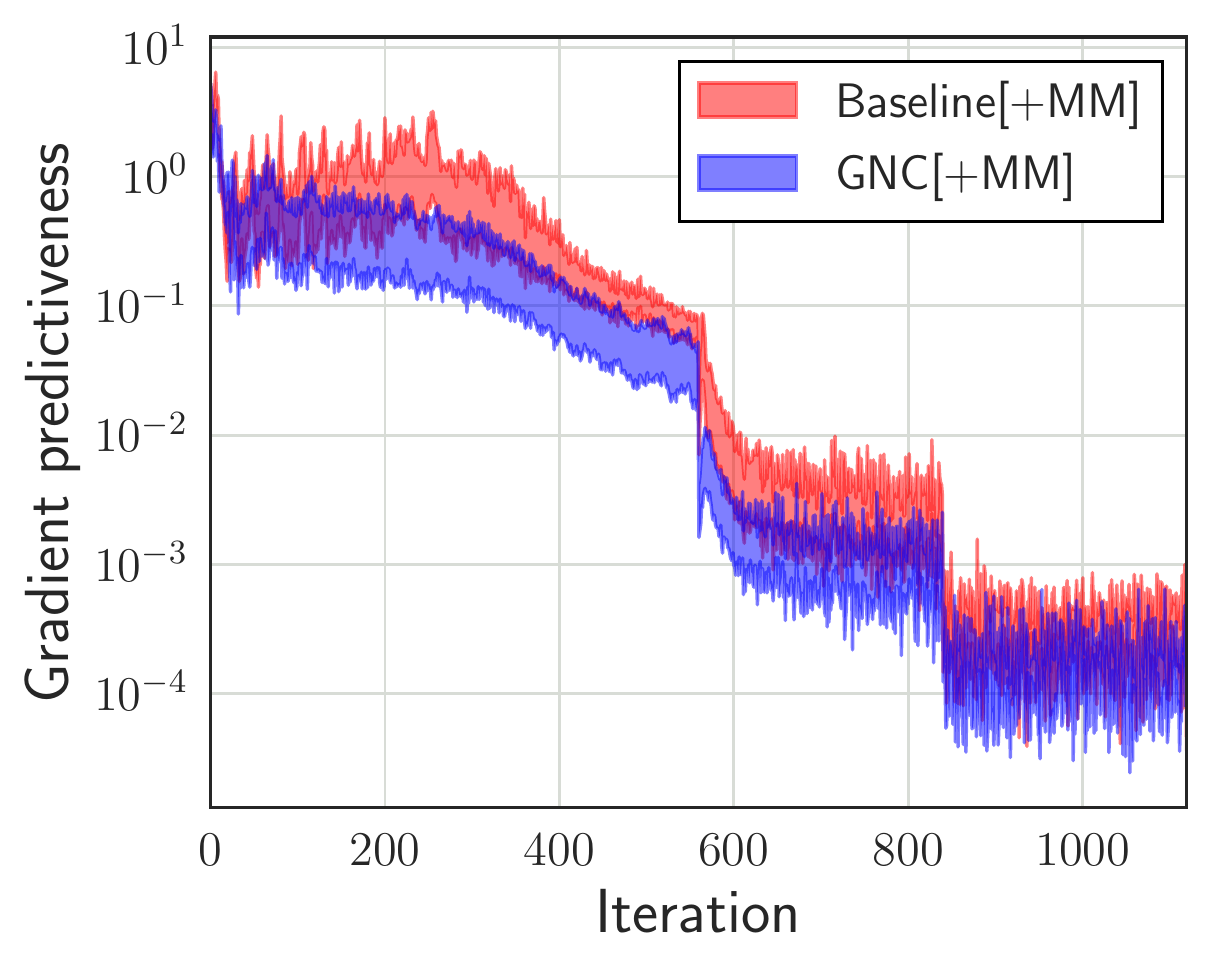}}
      \subfigure[Baseline + DA]{
        \def\subfigcapskip{0pt}
        \includegraphics[width=1.8in]{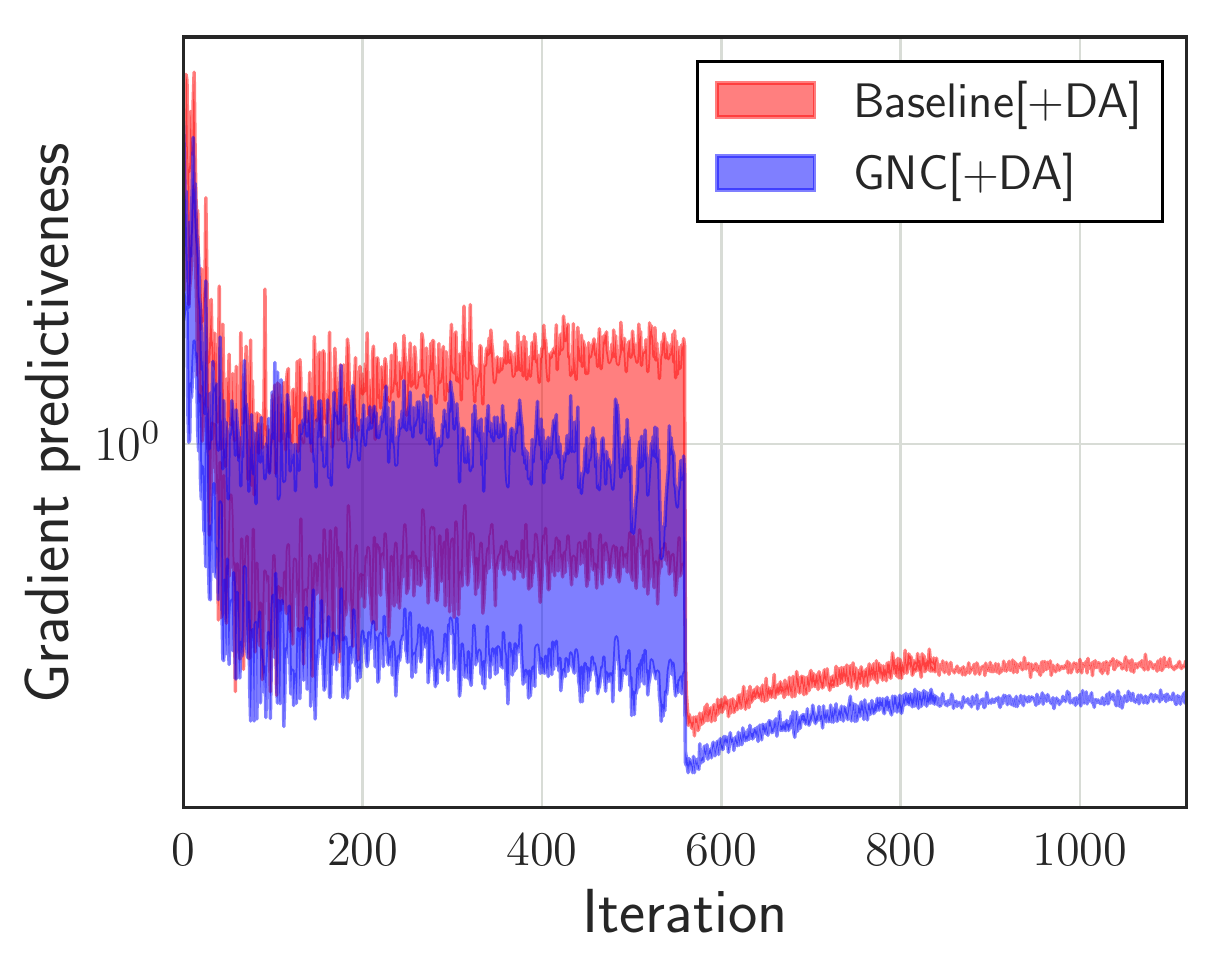}}
    \end{center}
  \end{minipage}
  \begin{minipage}{0.33\hsize}
    \begin{center}
      \subfigure[Baseline]{
        \def\subfigcapskip{0pt}
        \includegraphics[width=1.8in]{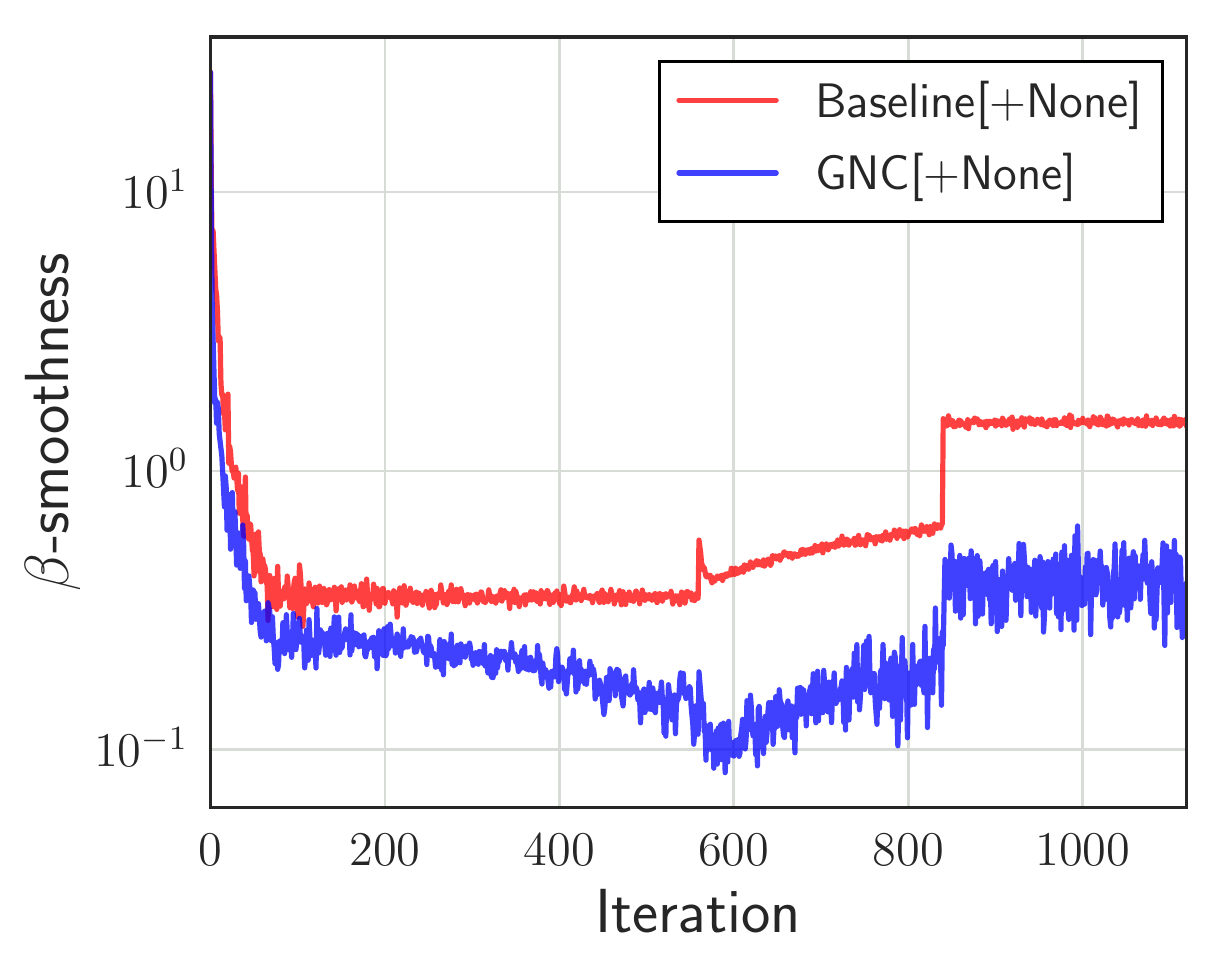}}
      \subfigure[Baseline + WD]{
        \def\subfigcapskip{0pt}
        \includegraphics[width=1.8in]{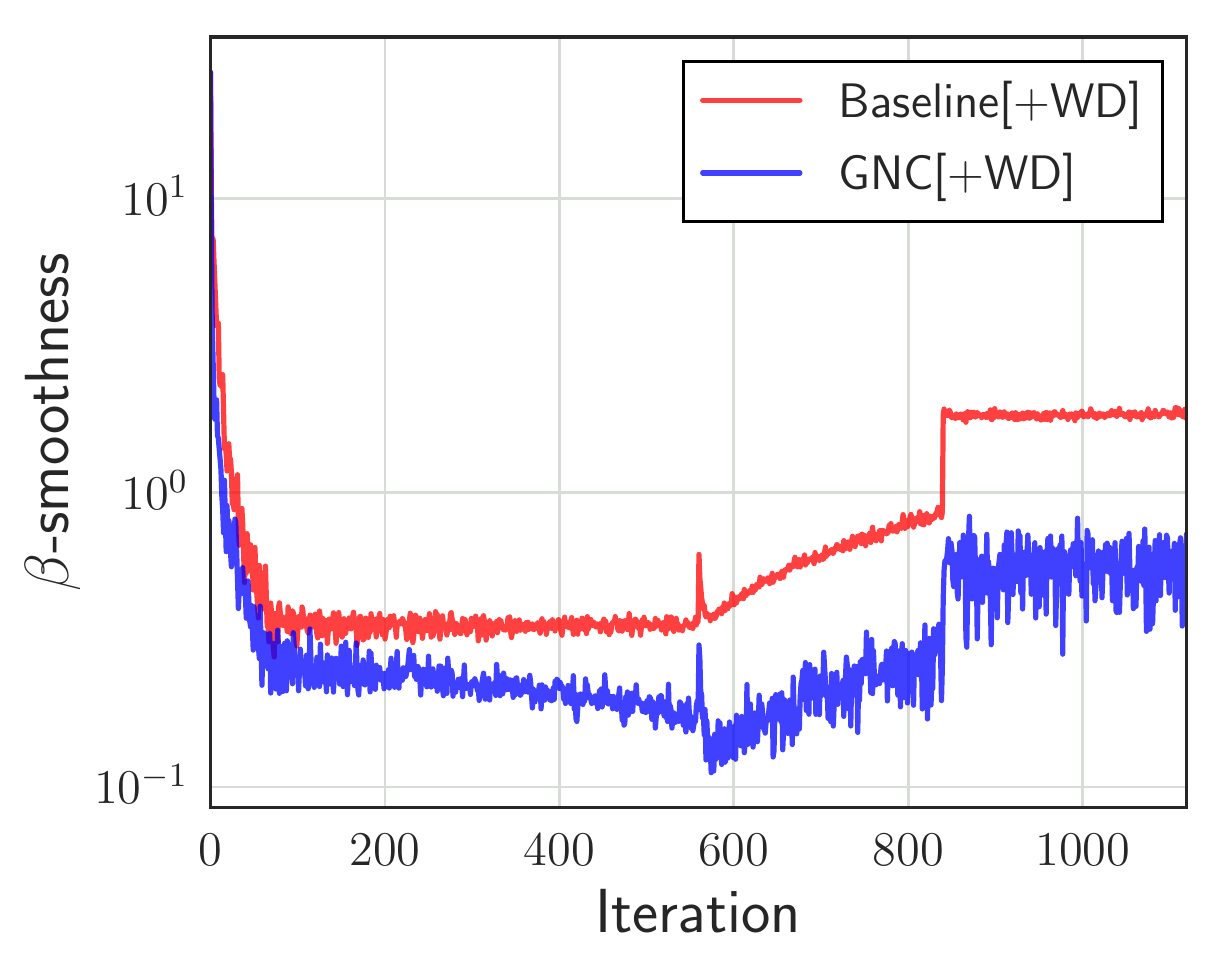}}
      \subfigure[Baseline + MM]{
        \def\subfigcapskip{0pt}
        \includegraphics[width=1.8in]{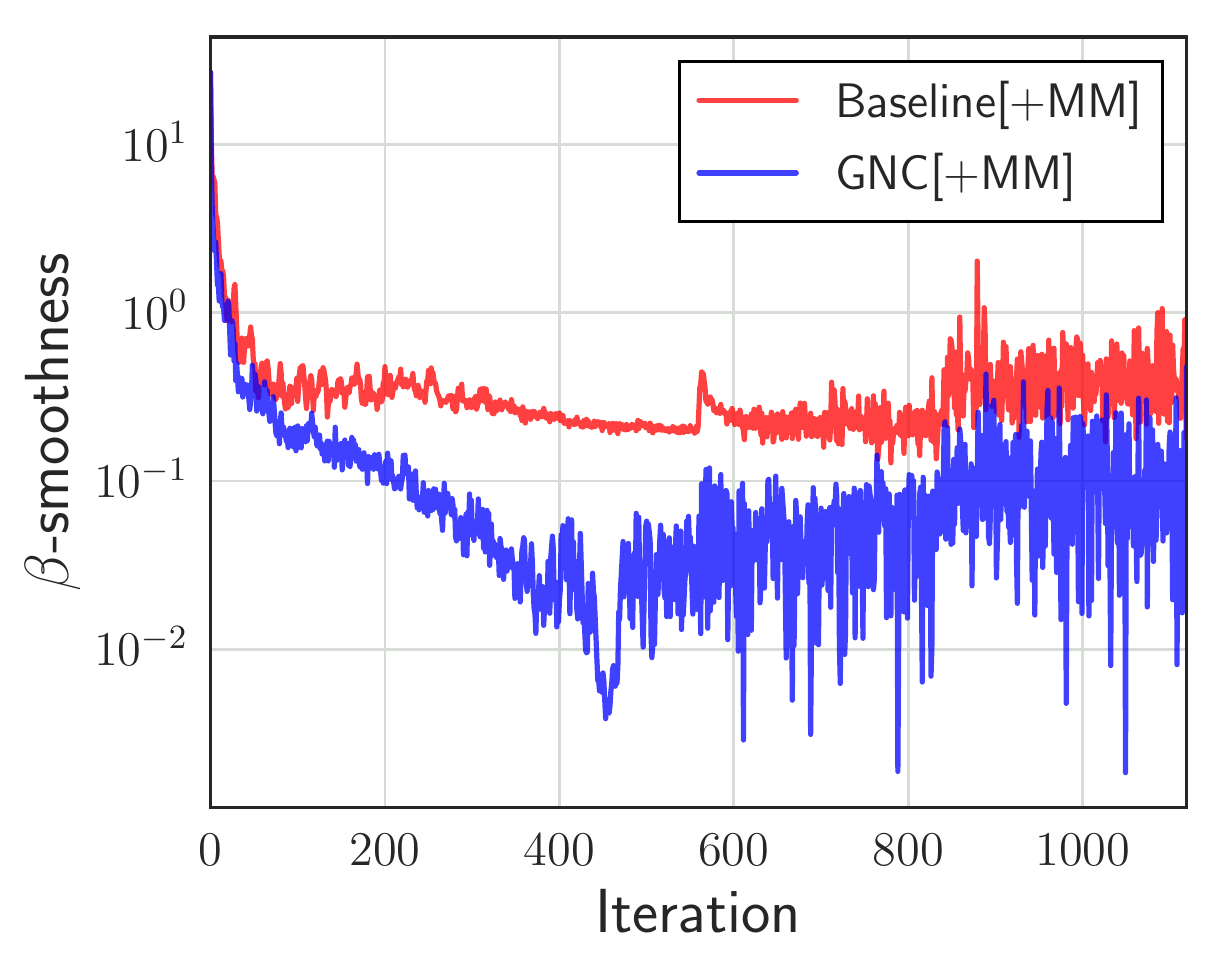}}
      \subfigure[Baseline + DA]{
        \def\subfigcapskip{0pt}
        \includegraphics[width=1.8in]{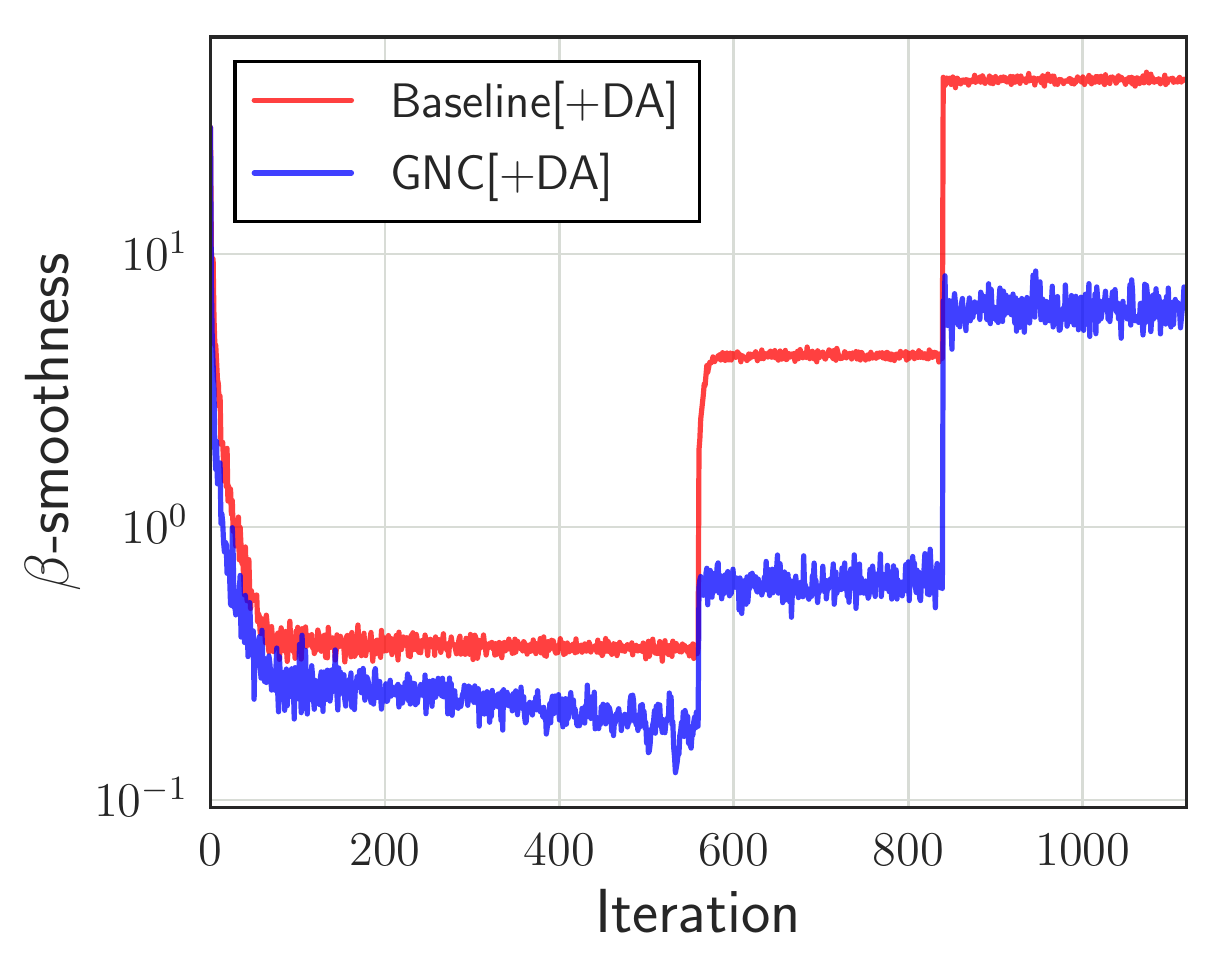}}
    \end{center}
  \end{minipage}
  \vskip -0.15in
  \caption{
    Smoothness of the loss function with and without GNC for the varieties in \ref{subsec:gnc_robustness}.
    Calculations of the three indicators---
    loss landscape, gradient predictiveness, and $\beta$-smoothness---
    are the same as in Fig.~\ref{fig:cifar100_lipschizness}.
    All indicators show clear improvement by GNC.
  }
  \label{fig:cifar100_rubust_lipschizness}
  \vskip -0.15in
\end{figure}

\begin{figure}[h]
  \begin{center}
  \subfigure[batch size of 32,768]{
    \def\subfigcapskip{0pt}
    \includegraphics[width=3in]{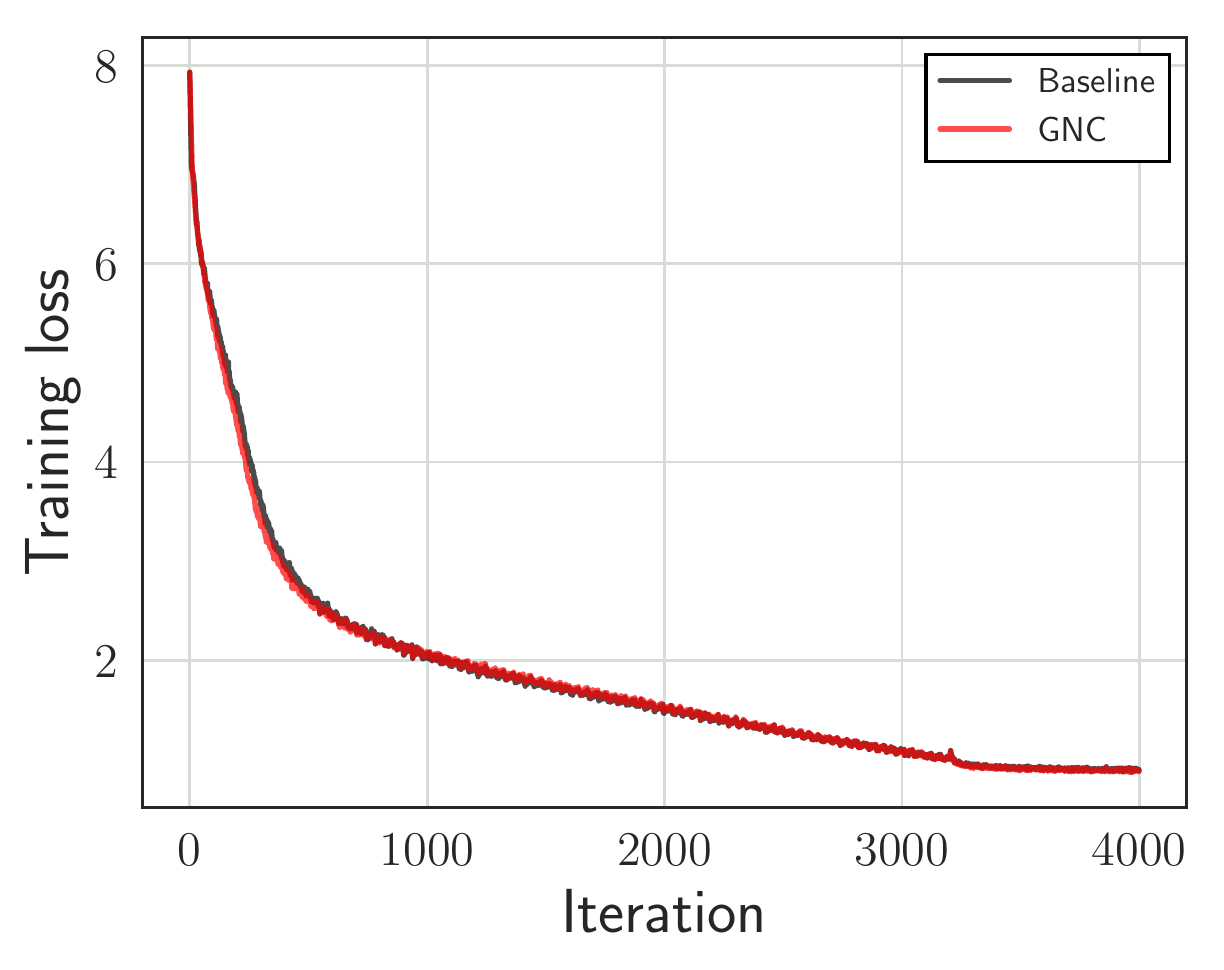}}
  \subfigure[batch size of 131,072]{
    \def\subfigcapskip{0pt}
    \includegraphics[width=3in]{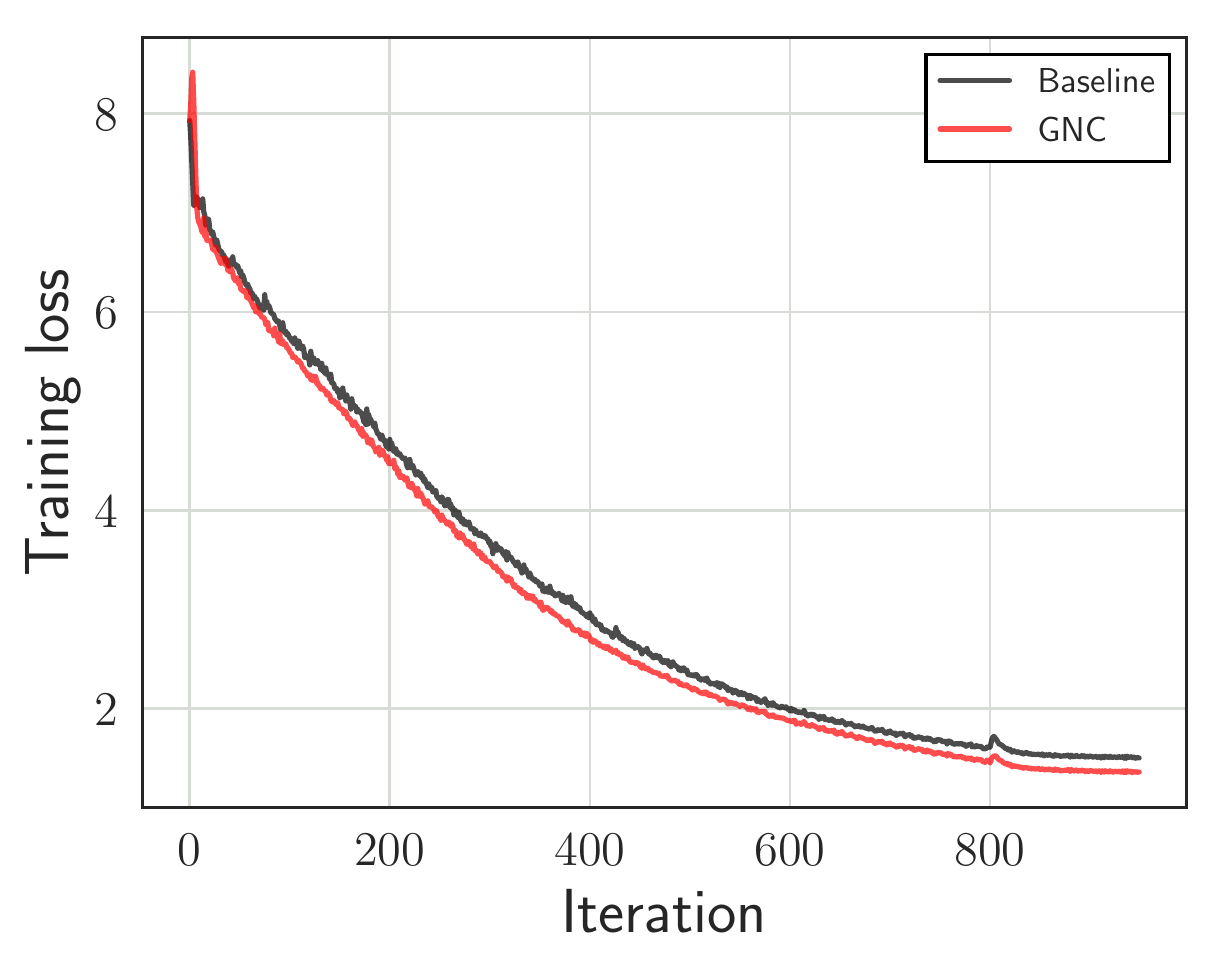}}
  \caption{
    ImageNet training loss curve with batch sizes of 32,768 and 131,072.
    In both cases, GNC retains smoother training loss than does the baseline (no noise) through the training trajectory,
    though by only a little in the batch size of 32,768.
    In the batch size of 131,072, the loss curve still decayed at the end of the curve, so
    switching to RNC around the 800th iterations might be premature, as shown in Table~\ref{app:table_rawdata_val_acc}.
  }
  \label{fig:fig_imagenet_b120k_train_loss}
  \end{center}
\end{figure}

\begin{figure}[h]
  \begin{center}
    \subfigure[batch size of 4,096]{
      \def\subfigcapskip{0pt}
      \includegraphics[width=3in]{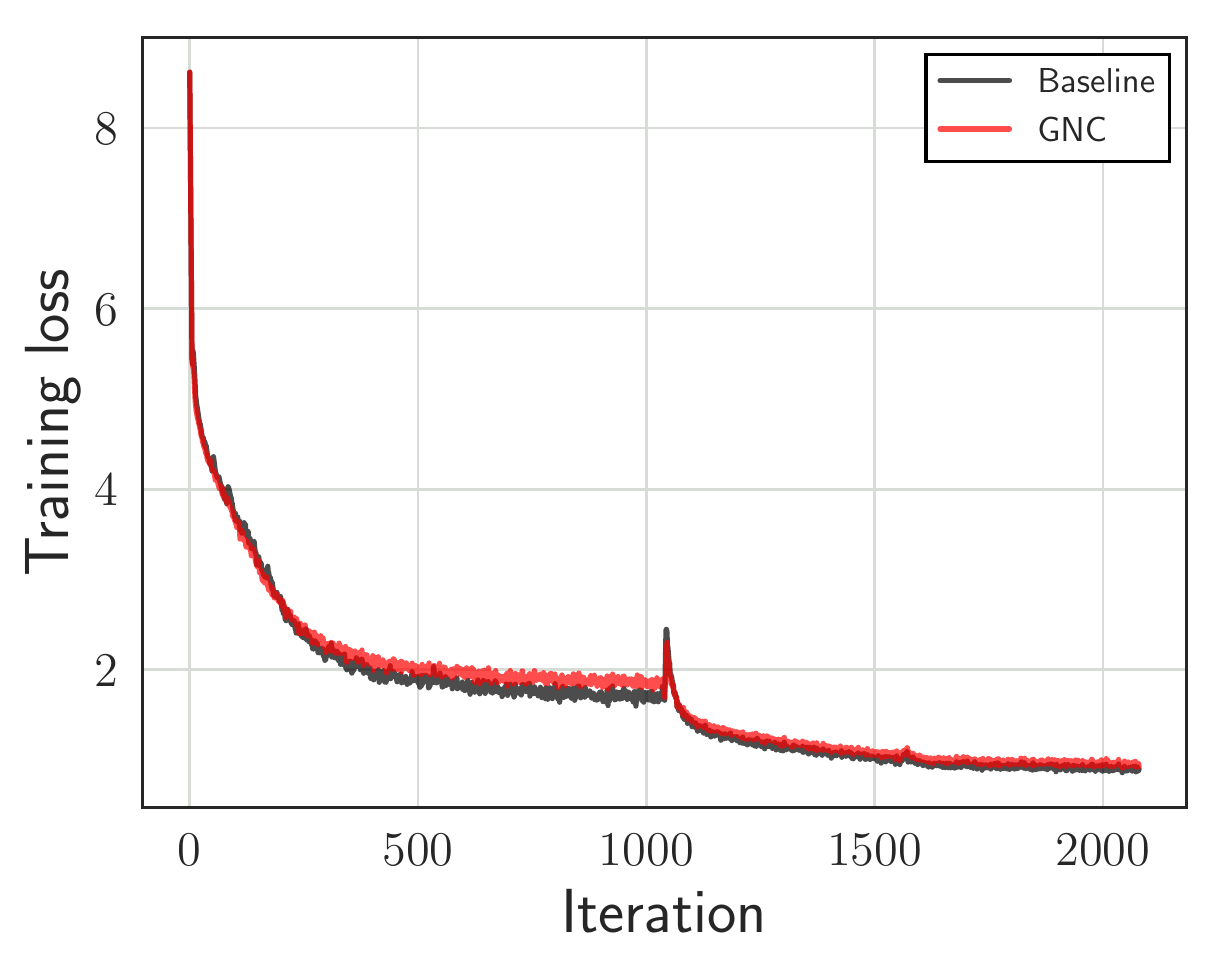}}
    \subfigure[batch size of 8,192]{
      \def\subfigcapskip{0pt}
      \includegraphics[width=3in]{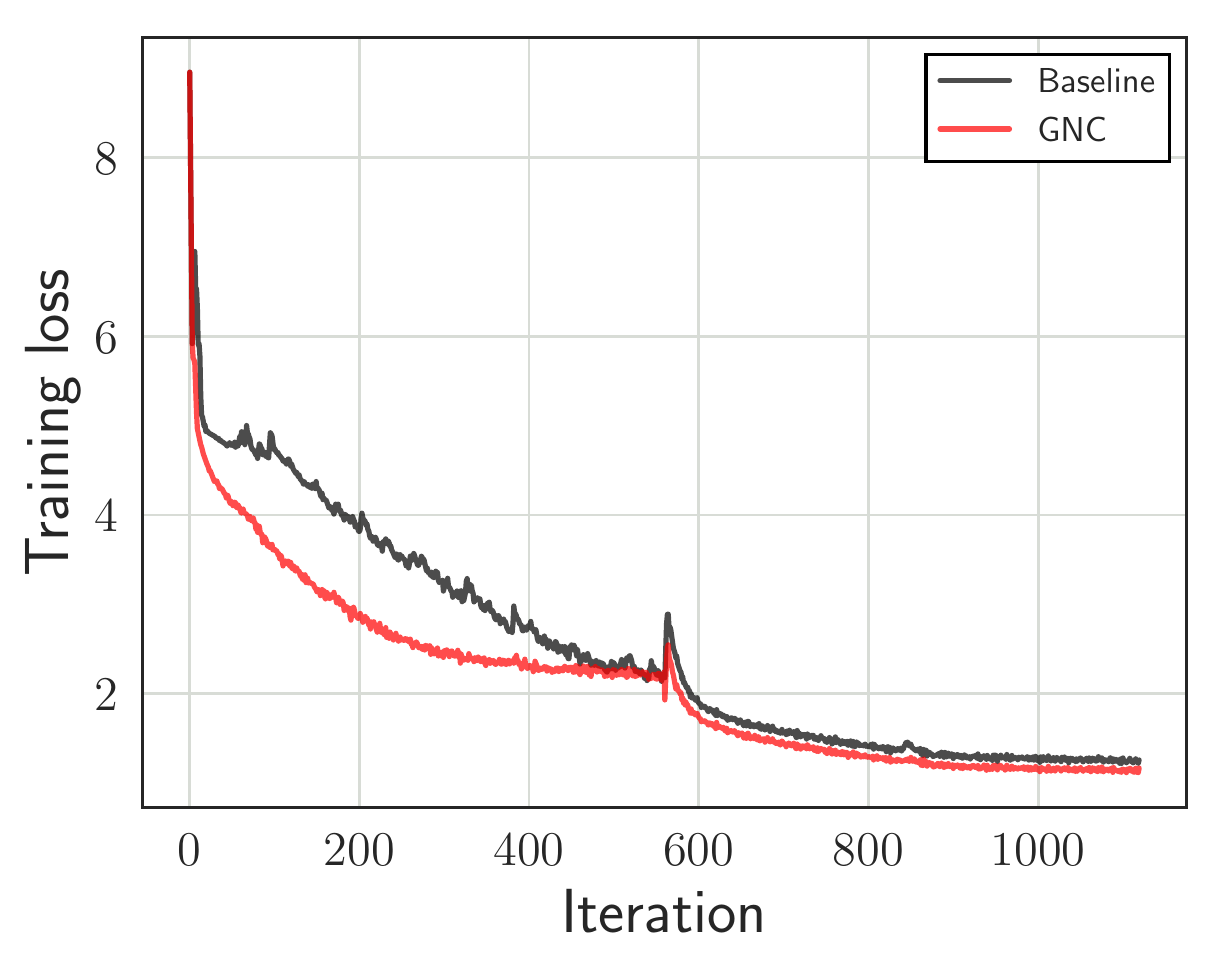}}
  \caption{
    CIFAR-100 training loss curve with batch sizes of 4,096 and 8,192.
    Besides the smoothing effect of GNC as seen in Sec.~\ref{subsec:gnc_smoothing_evidence},
    GNC also smooths the training curve.
  }
  \label{fig:fig_cifar100_b8k_train_loss}
  \end{center}
\end{figure}

\clearpage

\subsection{Raw Data for Validation Accuracies}
\label{subsec:raw_data}

\begin{table}[h]
\caption{
  Raw data for the validation accuracies in Table.~\ref{table_accuracies2}.
  As mentioned in Sec.~\ref{empirical_analysis}, for each Test ID, different methods
  (baseline, RNC, GNC, and GNCtoRNC) used the same initial random weight,
  random mini-batch sampling, and deterministic mode for the GPU.
  CIFAR-10 and 100 results do not include randomness and are completely reproducible,
  due to their being performed in a single node.
  ImageNet results include floating-point randomness,
  due to performing AllReduce, which is a collective communication operation among multi-nodes.
  In ImageNet with a batch size of 32,768, we achieved best accuracy (76.29\%; denoted by an asterisk) by GNC,
  though GNCtoRNC results are more stable. This might be due to the predictiveness drop of the full gradient,
  as described in Sec.~\ref{subsec:similarity_fg_lb}.
  Fig.~\ref{fig:imagenet_loss_distribution} also shows that GNC does not effectively vary losses
  in the latter part of the training process.
  In ImageNet with a batch size of 131,072, GNC achieved the best score.
  The reason for degradation with GNCtoRNC is shown in Fig.~\ref{fig:fig_imagenet_b120k_train_loss}.
  For CIFAR-10, GNC or GNCtoRNC achieved best scores.
  In particular, note the stability in the batch size of 8,192 by GNC or GNCtoRNC.
  The baseline method might be unstable due to the larger batch size.
  For CIFAR-100 there is a clear performance gain by GNC or GNCtoRNC.
  }
\label{app:table_rawdata_val_acc}
\begin{center}
\begin{small}
\begin{tabular}{lcccccc}
\toprule
	Task & Batch size & Test ID & Baseline & RNC & GNC & GNCtoRNC \\
\midrule
		  &         &0 & 75.91 & 75.67       & {\bf 76.29}$^{*}$ & 75.95 \\
		  &         &1 & 75.98 & 75.86       & 75.87            & {\bf 76.09} \\
	    & 32,768  &2 & 75.89 & 75.87       & {\bf 76.13}      & {\bf 76.13} \\
		  &         &3 & 75.92 & 76.02       & 75.96            & {\bf 76.06} \\
		  &         &4 & 75.74 & 75.86       & 75.88            & {\bf 76.04} \\ \cline{2-7}
	ImageNet  &   &0 & 66.14 & 65.75       & {\bf 68.81}      & 68.80 \\
		  &         &1 & 65.99 & 64.96       & {\bf 68.11}      & 67.57 \\
	    & 131,072 &2 & 65.22 & 64.27       & {\bf 68.34}      & 68.32 \\
		  &         &3 & 65.24 & 66.29       & {\bf 68.78}      & 68.17 \\
		  &         &4 & 65.27 & 64.44       & {\bf 67.92}      & 67.80 \\ \hline
		  &         &0 & 93.73 & 93.60       & 94.32            & {\bf 94.48} \\
		  &         &1 & 93.47 & {\bf 93.76} & 92.72            & 93.00 \\
	    & 4,096   &2 & 93.12 & 93.31       & {\bf 93.97}      & 94.10 \\
		  &         &3 & 93.74 & 93.81       & 94.17            & {\bf 94.42} \\
		  &         &4 & 93.34 & 94.00       & 93.92            & {\bf 94.01} \\ \cline{2-7}
	CIFAR-10  &   &0 & 66.84 & 72.28       & {\bf 90.93}      & 90.73 \\
		  &         &1 & 52.43 & 43.93       & 92.14            & {\bf 92.17} \\
	    & 8,192   &2 & 50.92 & 78.04       & 90.84            & {\bf 91.02} \\
		  &         &3 & 49.34 & 39.36       & 92.02            & {\bf 92.11} \\
		  &         &4 & 53.00 & 83.96       & {\bf 89.06}      & 88.39 \\ \hline
		  &         &0 & 72.93 & 73.35       & 73.32            & {\bf 73.95} \\
		  &         &1 & 72.98 & 73.24       & 73.01            & {\bf 74.07} \\
	    & 4,096   &2 & 72.34 & 73.23       & 72.36            & {\bf 73.10} \\
		  &         &3 & 73.36 & 72.75       & 72.67            & {\bf 73.79} \\
		  &         &4 & 72.76 & 72.84       & 73.10            & {\bf 74.05} \\ \cline{2-7}
	CIFAR-100 &   &0 & 67.81 & 68.63       & 71.54            & {\bf 72.21} \\
		  &         &1 & 68.99 & 70.95       & 71.11            & {\bf 71.68} \\
	    & 8,192   &2 & 71.09 & 70.84       & 71.64            & {\bf 72.07} \\
		  &         &3 & 66.60 & 68.80       & 71.02            & {\bf 71.82} \\
		  &         &4 & 71.55 & 71.61       & 71.45            & {\bf 71.88} \\
\bottomrule
\end{tabular}
\end{small}
\end{center}
\end{table}

\clearpage

\subsection{Related Works for ImageNet Benchmarks}
\label{app:subsec:imagenet_benchmark}


\begin{table}[h]
\caption{
  Results of ImageNet on ResNet-50 training with large-batch sizes of 32,768
  (except for the case of \citet{Jia18}, which was 65,536).
  Under benchmark settings of 90 epochs and a batch size of 32,768,
  our method realizes state-of-the-art performance on validation accuracy.
  Note that \citet{Jia18} and \citet{Ying18} did not reveal the details of their experimental settings,
  so those results are beyond the scope of this paper.
  }
\label{app:imagenet_table_benchmark}
\begin{center}
\begin{small}
\begin{tabular}{lccc}
\toprule
&  Epochs & Batch size & Accuracy(\%) \\
\midrule
	\citet{Goyal17} & 90 & 32,768 & 72.45 \\
	\citet{Akiba17} & 90 & 32,768 & 74.94 \\
	\citet{Condreanu17} & 100 & 32,768 & 75.31 \\
	\citet{You18}& 90 & 32,768 & 75.4 \\
	{\it \citet{Jia18}} & {\it 90} & {\it 65,536} & {\it 76.2}$^{*}$ \\
	{\it \citet{Ying18}} & {\it 90} & {\it 32,768} & {\it 76.4}$^{*}$ \\
	Ours & 90 & 32,768 & {\bf 76.29} \\
\bottomrule
\end{tabular}
\end{small}
\end{center}
\end{table}

\end{document}